\documentclass[10pt,twocolumn,letterpaper]{article}

\usepackage[pagenumbers]{cvpr} % To force page numbers, e.g. for an arXiv version

\usepackage{graphicx}
\usepackage{amsmath}
\usepackage{amssymb}
\usepackage{booktabs}

\makeatletter
\@namedef{ver@everyshi.sty}{}
\makeatother
\usepackage{tikz}
\usepackage[accsupp]{axessibility}
\usepackage{comment}
\usepackage[font=small]{caption}
\usepackage{colortbl}
\usepackage{subcaption}
\usepackage{bbm}
\usepackage{color}

\usepackage[pagebackref,breaklinks,colorlinks]{hyperref}

\usepackage[capitalize]{cleveref}
\crefname{section}{Sec.}{Secs.}
\Crefname{section}{Section}{Sections}
\Crefname{table}{Table}{Tables}
\crefname{table}{Tab.}{Tabs.}

\usepackage{overpic}
\usepackage{enumitem} %< control spacing in itemize/enumerate/...
\usepackage{overpic} %< add raw math symbols to figures
\usepackage{color}
\usepackage{wrapfig}

\makeatletter
\DeclareRobustCommand\onedot{\futurelet\@let@token\@onedot}
\def\@onedot{\ifx\@let@token.\else.\null\fi\xspace}

\def\ie{\emph{i.e}\onedot}

\def\etal{\emph{et al}\onedot}
\makeatother

\definecolor{turquoise}{cmyk}{0.65,0,0.1,0.3}
\definecolor{purple}{rgb}{0.65,0,0.65}
\definecolor{dark_green}{rgb}{0, 0.5, 0}
\definecolor{orange}{rgb}{0.8, 0.6, 0.2}
\definecolor{red}{rgb}{0.8, 0.2, 0.2}
\definecolor{darkred}{rgb}{0.6, 0.1, 0.05}
\definecolor{blueish}{rgb}{0.0, 0.3, .6}
\definecolor{light_gray}{rgb}{0.7, 0.7, .7}
\definecolor{pink}{rgb}{1, 0, 1}
\definecolor{greyblue}{rgb}{0.25, 0.25, 1}

\usepackage{blindtext}

\renewcommand{\paragraph}[1]{\vspace{1em}\noindent\textbf{#1}.}

\newcommand{\eqn}[1]{Equation~\ref{#1}}

\definecolor{JonYellow}{rgb}{1,1, 0.6}
\definecolor{JonOrange}{rgb}{1, 0.8, 0.6}
\definecolor{JonRed}{rgb}{1, 0.6, 0.6}

\usepackage[ruled]{algorithm2e} % For algorithms

\SetAlFnt{\small}
\SetAlCapFnt{\small}
\SetAlCapNameFnt{\small}
\SetAlCapHSkip{0pt}

\usepackage{cancel}

\def\cvprPaperID{2231} % *** Enter the CVPR Paper ID here
\def\confName{CVPR}
\def\confYear{2023}

\begin{document}

\title{SunStage: Portrait Reconstruction and Relighting using the Sun as a Light Stage}
% \title{Supplementary Materials for \\ SunStage: Portrait Reconstruction and Relighting using the Sun as a Light Stage}

\author{Yifan Wang$^1$
\quad
Aleksander Holynski$^1$
\quad
Xiuming Zhang$^2$
\quad
\vspace{2mm}
Xuaner Zhang$^2$ \\
{$^1$University of Washington \qquad $^2$Adobe Inc.} \\
\href{https://grail.cs.washington.edu/projects/sunstage/}{sunstage.cs.washington.edu}
}

% for main paper
\twocolumn[{%
\renewcommand\twocolumn[1][]{#1}%
\begin{center}
\maketitle
\vspace{-1.5em}
\begin{tabular}{@{}c@{\hspace{1mm}}c@{\hspace{1mm}}c@{\hspace{1mm}}c@{\hspace{1mm}}c@{}}
\captionsetup{type=figure}
\begin{subfigure}[b]{0.19\linewidth}
% \includegraphics[width=\linewidth]{fig/teaser/dan_input.png} %
% \end{subfigure}
% \begin{subfigure}[b]{0.19\linewidth}
% \includegraphics[width=\linewidth]{fig/teaser/dan_relight.png} %
% \end{subfigure}
% \begin{subfigure}[b]{0.19\linewidth}
% \includegraphics[width=\linewidth]{fig/teaser/dan_edit_albedo.png} %
% \end{subfigure}
% \begin{subfigure}[b]{0.19\linewidth}
% \includegraphics[width=\linewidth]{fig/teaser/dan_shiny.png} %
% \end{subfigure}
% \begin{subfigure}[b]{0.19\linewidth}
% \includegraphics[width=\linewidth]{fig/teaser/dan_envmap_inset.png} %

\includegraphics[width=\linewidth]{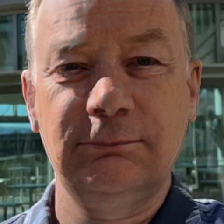} %
\end{subfigure}
\begin{subfigure}[b]{0.19\linewidth}
\includegraphics[width=\linewidth]{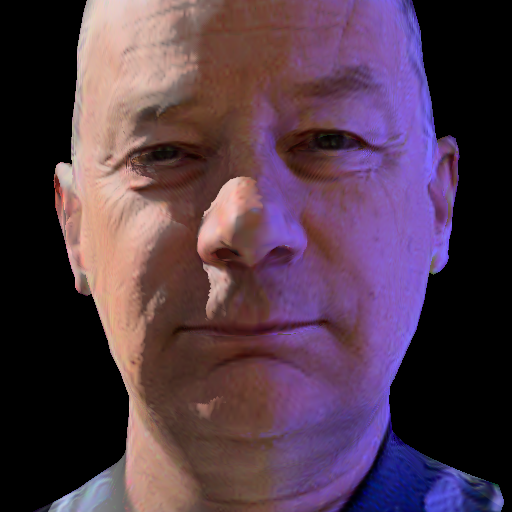} %
\end{subfigure}
\begin{subfigure}[b]{0.19\linewidth}
\includegraphics[width=\linewidth]{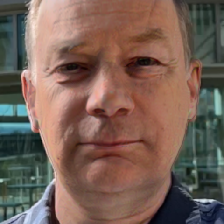} %
\end{subfigure}
\begin{subfigure}[b]{0.19\linewidth}
\includegraphics[width=\linewidth]{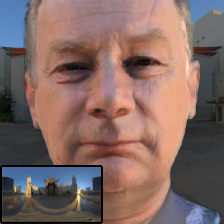} %
\end{subfigure}
\begin{subfigure}[b]{0.19\linewidth}
\includegraphics[width=\linewidth]{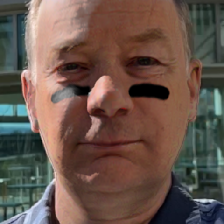} %
\end{subfigure}\\
\captionsetup{type=figure}
%\begin{subfigure}[b]{0.19\linewidth}
%\includegraphics[width=\linewidth]{fig/teaser/florian_shiny.png} %
%\end{subfigure}\\
\begin{subfigure}[b]{0.19\linewidth}
\includegraphics[width=\linewidth]{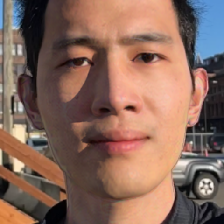} %
\caption{Reconstruction}
\end{subfigure}
\begin{subfigure}[b]{0.19\linewidth}
\includegraphics[width=\linewidth]{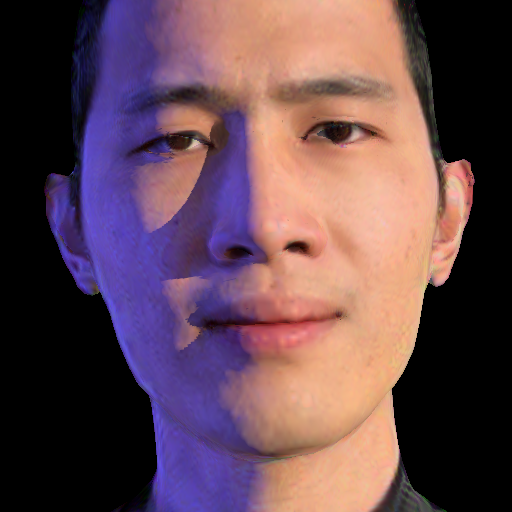} %
\caption{Edit lighting}
\end{subfigure}
\begin{subfigure}[b]{0.19\linewidth}
\includegraphics[width=\linewidth]{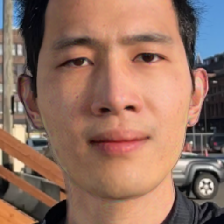} %
\caption{Soften shadows}
\end{subfigure}
\begin{subfigure}[b]{0.19\linewidth}
\includegraphics[width=\linewidth]{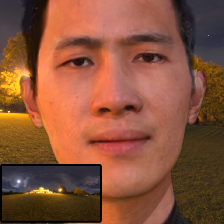} %
\caption{Swap lighting}
\end{subfigure}
\begin{subfigure}[b]{0.19\linewidth}
\includegraphics[width=\linewidth]{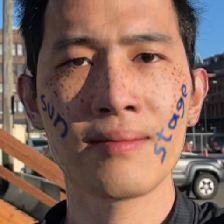} %
\caption{Edit texture}
\end{subfigure}
\vspace{-1em}
%\begin{subfigure}[b]{0.19\linewidth}
%\includegraphics[width=\linewidth]{fig/teaser/yifan_shiny.png} %
%\caption{Edit materials}
%\end{subfigure}
\end{tabular}
\addtocounter{figure}{-2}
\captionof{figure}{Given a selfie video rotating under the sun, SunStage reconstructs geometry, material, camera pose, and lighting information. This recovered information can be used to (a) realistically re-render the input images, (b) modify the lighting conditions by adding / removing lights, (c) soften harsh shadows by changing the size of the reconstructed light sources (d) render the person in an entirely new environment, and (e) edit the albedo or material properties to add freckles, makeup, or stickers that realistically interact with scene lighting.}
%edit the camera viewpoint or focal length.}
\label{fig:teaser}
% \vspace{-5mm}
\end{center}%
% \end{figure*}
}]

\begin{abstract}
%Outdoor portrait photographs are often marred by the harsh shadows cast under direct sunlight.
%To resolve this, one can use post-capture lighting manipulation techniques, but these methods either require complex hardware (e.g., a light stage) to capture each individual, or rely on image-based priors and thus fail to reconstruct many of the subtle facial details that vary from person to person.
%In this paper, we present SunStage, a system for accurate, individually-tailored, and lightweight reconstruction of facial geometry and reflectance that can be used for general portrait relighting with cast shadows.
%Our method only requires the user to capture a selfie video outdoors, rotating in place, and uses the varying angles between the sun and the face as constraints in the joint reconstruction of facial geometry, reflectance properties, and lighting parameters.
%Aside from relighting, we show that our reconstruction can be used for applications like reflectance editing and view synthesis.

\vspace{-1em}A light stage %are well-calibrated collections of cameras and lights that can capture a human subject under varying illumination and viewpoint. 
uses a series of calibrated cameras and lights to capture a subject's facial appearance under varying illumination and viewpoint. 
%series of images captured under synchronized cameras and lights. 
This captured information is crucial for facial reconstruction and relighting. Unfortunately, light stages are often inaccessible: they are expensive and require significant technical expertise for construction and operation. In this paper, we present SunStage: a lightweight alternative to a light stage that captures comparable data using only a smartphone camera and the sun. Our method only requires the user to capture a selfie video outdoors, rotating in place, and uses the varying angles between the sun and the face as guidance in joint reconstruction of facial geometry, reflectance, camera pose, and lighting parameters. Despite the in-the-wild un-calibrated setting, our approach is able to reconstruct detailed facial appearance and geometry, enabling compelling effects such as relighting, novel view synthesis, and reflectance editing.\end{abstract}
\vspace{-1.8em}\section{Introduction}
\label{sec:intro}
%
% It is challenging to capture a visually pleasing portrait photograph outdoors.
% Many desired qualities of a portrait photograph, such as smooth and even lighting, are difficult to achieve under direct sunlight, as the sun casts harsh shadows and accentuates unflattering details like wrinkles (e.g., in Figure~\ref{fig:teaser}a).
% Professional portrait photographers have long perfected approaches for combating these difficult lighting conditions by adding lights in order to produce softer and more even lighting~\cite{hunter2021light}.
% These approaches typically rely on elaborate lighting setups consisting of scrims (cloth diffusers), reflectors, flashes, and bounce cards~\cite{grey2014master}.
% Unfortunately, in most everyday scenarios, instead of a professional photographer or complex lighting kits, casual users only have access to a single portable camera (e.g., a smartphone).
% In these cases, one must rely on alternative solutions for adjusting face lighting, such as recent methods that enable virtual manipulation of the lighting post-capture~\cite{zhou2019deep,zhang2021neural,zhang2020portrait,sun2019single,pandey2021total}.
A light stage~\cite{debevec2000acquiring} acquires the shape and material properties of a face in high detail using a series of images captured under synchronized cameras and lights. This captured information can be used to synthesize novel images of the subject under arbitrary lighting conditions or from arbitrary viewpoints. This process enables a number of visual effects, such as creating digital replicas of actors that can be used in movies~\cite{alexander2010digital} or high-quality postproduction relighting~\cite{wenger2005performance}.

In many cases, however, it is often infeasible to get access to a light stage for capturing a particular subject, because light stages are not easy to find: they are expensive and require significant technical expertise (often teams of people) to build and operate. In these cases, hope is not lost --- one can turn to methods that are \emph{trained} on light stage data, with the intention of generalizing to new subjects. These methods do not require the subject to be captured by a light stage but instead use a machine learning model trained on a collection of previously acquired light stage captures to enable the same applications as a light stage, but from only one or several images of a new subject~\cite{pandey2021total,sun2019single,zhang2020portrait,booth20163d,li2017learning,sengupta2018sfsnet,zhou2019deep}.
%
%either because of is often unavailable or unattainable for capturing a particular subject. In these (very common) cases, one can turn to alternative approaches that 
%
Unfortunately, these methods have difficulty faithfully reproducing and editing the appearance of new subjects, as they lack much of the signal necessary to resolve the ambiguities of single-view reconstruction, i.e., a single image of a face can be reasonably explained by different combinations of geometry, illumination, and reflectance. % and to reconstruct the spatially-varying reflectance characteristics that often vary from subject to subject.
%Additionally, these methods do not explicitly estimate geometry and thus cannot simulate many of the subtle but necessary light transport effects seen on faces, such as cast shadows (e.g., those cast by the nose over the cheek). 
%Some methods do estimate geometry, and rely on face shape priors from morphable face models~\cite{}, which in theory can enable physically-correct editing~\cite{sengupta2018sfsnet,zhou2019deep}.
%In practice, however, these methods have difficulty resolving the ambiguities of single-view reconstruction, \ie, a single image of a face can be reasonably explained by different combinations of geometry, illumination, and reflectance. 

In this paper, we propose an intermediate solution --- one that allows for personalized, high-quality capture of a given subject, but without the need for expensive, calibrated capture equipment. 
%these limitations by proposing a method to enable high-quality, physically-grounded portrait relighting without the need for additional hardware (or even another person).
Our method, which we dub SunStage, uses only a handheld smartphone camera and the sun to simulate a minimalist light stage, enabling the reconstruction of individually-tailored geometry and reflectance without specialized equipment.
Our capture setup only requires the user to hold the camera at arm's length and rotate in place, allowing the face to be observed under varying angles of incident sunlight, which causes specular highlights to move and shadows to swing across the face. This provides strong signals for the reconstruction of facial geometry and spatially-varying reflectance properties.
The reconstructed face and scene parameters estimated by our system can be used to realistically render the subject in new, unseen lighting conditions --- even with complex details like self-occluding cast shadows, which are typically missing in purely image-based relighting techniques, \ie, those that do not explicitly model geometry. In addition to relighting, we also show applications in view synthesis, correcting facial perspective distortion, and editing skin reflectance.

Our contributions include: (1) a novel capture technique for personalized facial scanning without custom equipment, (2) a system for optimization and disentanglement of scene parameters (geometry, materials, lighting, and camera poses) from an unaligned, handheld video, and (3) multiple portrait editing applications that produce photorealistic results, using as input only a single selfie video.

\section{Related works}
\label{sec:related}

%Our work is related to face modeling, portrait relighting, appearance decomposition, as well as general object relighting. 

\paragraph{Face modeling}
Extensive research has been devoted to the modeling of human faces, leading to various 3D morphable models (3DMMs) \cite{blanz1999morphable,cao2013facewarehouse,brunton2014multilinear,bolkart2015groupwise,booth20163d,dai20173d,li2017learning,ranjan2018generating,ploumpis2019combining,pighin2006synthesizing,tran2018nonlinear,bi2021avatar}.
These models are parametric (maybe in the form of neural networks \cite{ranjan2018generating}), allowing one to express variations compactly with a vector.
They also encode strong priors learned from real scans.
The groundbreaking face 3DMM is that of Blanz and Vetter \cite{blanz1999morphable} containing models for shape, expression, and appearance (the Phong model).
Also influential is the FLAME model~\cite{li2017learning} that uses vertex-based Linear Blend Skinning (LBS).
FLAME is described by a mapping from shape, pose, and expression vectors to a list of vertices.
We refer the reader to the survey by Egger \etal \cite{egger20203d} for different face morphable models.

Such parametric face models provide a low-dimensional space for optimization or learning algorithms.
DECA~\cite{feng2021learning} uses the FLAME model to estimate detailed facial geometry (and albedo) from single images, by predicting additional displacement maps and adding them to the estimated FLAME models. More recently, NextFace~\cite{dib2021towards} employs the 3DMM geometry and albedo priors to learn an albedo residual that captures more facial details.
% We use DECA that is based on FLAME as our shape initialization.

Without modeling 3D face geometry, researchers have also achieved photorealistic synthesis of portrait images using generative models and large-scale high-quality image datasets \cite{karras2019style,karras2020analyzing}. 
% \yifan{Hou \etal~\cite{hou2022face} leveraged depth maps to model the facial geometry, and can synthesize geometrically consistent hard shadows via a differentiable ray tracing algorithm.}

\paragraph{Light stage capture}
The light stage as described in Debevec \etal, achieves impressive portrait reconstruction and relighting by capturing a series of images of the face under varying illumination~\cite{debevec2000acquiring}. Subsequent work made this process faster, more efficient, and explored different types of illuminants \cite{meka2019deep, ghosh2011multiview, fyffe2015single}.
%Along this line of using a light stage, Sun \etal developed a single-image portrait relighting network trained on ground-truth relit images generated by combining the OLAT images captured with a light stage \cite{}.
%The follow-up work by Sun \etal studies high-frequency, continuous portrait relighting beyond the physical sampling patterns of a light stage \cite{}.

Given that a light stage is not always accessible, a number of methods have been proposed to achieve similar outputs from a single (or few) input portrait images \cite{sun2019single, zhang2020portrait,zhang2021neuralvideo, pandey2021total,sun2020light,yeh2022learning, nestmeyer2020learning,hou2022face,hou2021towards}. These method rely on a dataset of light stage captures or synthetic examples as training data.  

%\paragraph{Facial relighting}
%With a setup such as a light stage, Debevec \etal achieve impressive portrait relighting by directly acquiring the reflectance field of a human face (\ie, capturing OLAT images) \cite{debevec2000acquiring}.
%Because direct acquisition of the reflectance field takes several seconds, researchers have also explored using structured illumination such as color gradient illumination to support relighting of a dynamic face \cite{meka2019deep}.
%Along this line of using a light stage, Sun \etal developed a single-image portrait relighting network trained on ground-truth relit images generated by combining the OLAT images captured with a light stage \cite{sun2019single}.
%The follow-up work by Sun \etal studies high-frequency, continuous portrait relighting beyond the physical sampling patterns of a light stage \cite{sun2020light}.
%With ground-truth supervision generated with a light stage, Total Relighting learns to predict an alpha matte and a relit portrait that together achieves lighting-aware background replacement \cite{pandey2021total}. Neural Light Transport (NLT) was developed to interpolate the light transport function given the sparse samples by a light stage to support relighting and viewpoint change \cite{zhang2021neural}. Relying on a light stage, Zhang \etal use an image-to-image network to perform video portrait relighting in real time \cite{zhang2021neuralvideo}. 

Our setup can be thought of as a ``minimalist light stage'' formed by just the sun and a rotating camera, without requiring the high construction and maintenance costs of building a light stage. This parameterization of a sun and skylight model has been shown to be effective in photometric stereo~\cite{holdgeoffroy-3dv-15,jung2015one} and scene factorization~\cite{Sunkavalli:2007:FTV,liu2020learning}.
In a similar spirit, Calian \etal~\cite{calian2018faces} focus on lighting estimation using faces as ``light probes''. Sengupta \etal propose to circumvent the need for a complicated light stage by recording the facial appearance responses to varying contents displayed on a desk monitor, and then perform portrait relighting \cite{sengupta2021light}.
Sevastopolsky \etal also attempt to simplify the capture setup from a light stage to a mobile phone camera with a co-located flash \cite{sevastopolsky2020relightable}. Unlike our work, which is physically-based, their approaches use neural rendering, and therefore have less direct control over lighting, material, and scene parameters.

%For photographic use cases, the user often prefers to \emph{enhance} rather than \emph{replace} the lighting condition at the time of capture. Towards this end, Zhang \etal proposed to learn a portrait shadow manipulation network from pairs of real photos and their corresponding ``shadowed versions'' generated synthetically in a careful way \cite{zhang2020portrait}.
%Using SPADE \cite{park2019SPADE} to ``inject'' lighting as a style into their network, Han \etal employ 3DMMs to provide 3D guidance for portrait lighting enhancement \cite{han2021deep}. However, these works cannot synthesize accurate cast shadows, which is crucial for portrait relighting.

\begin{figure*}[t!]
\begin{center}
\includegraphics[width=1.00\linewidth]{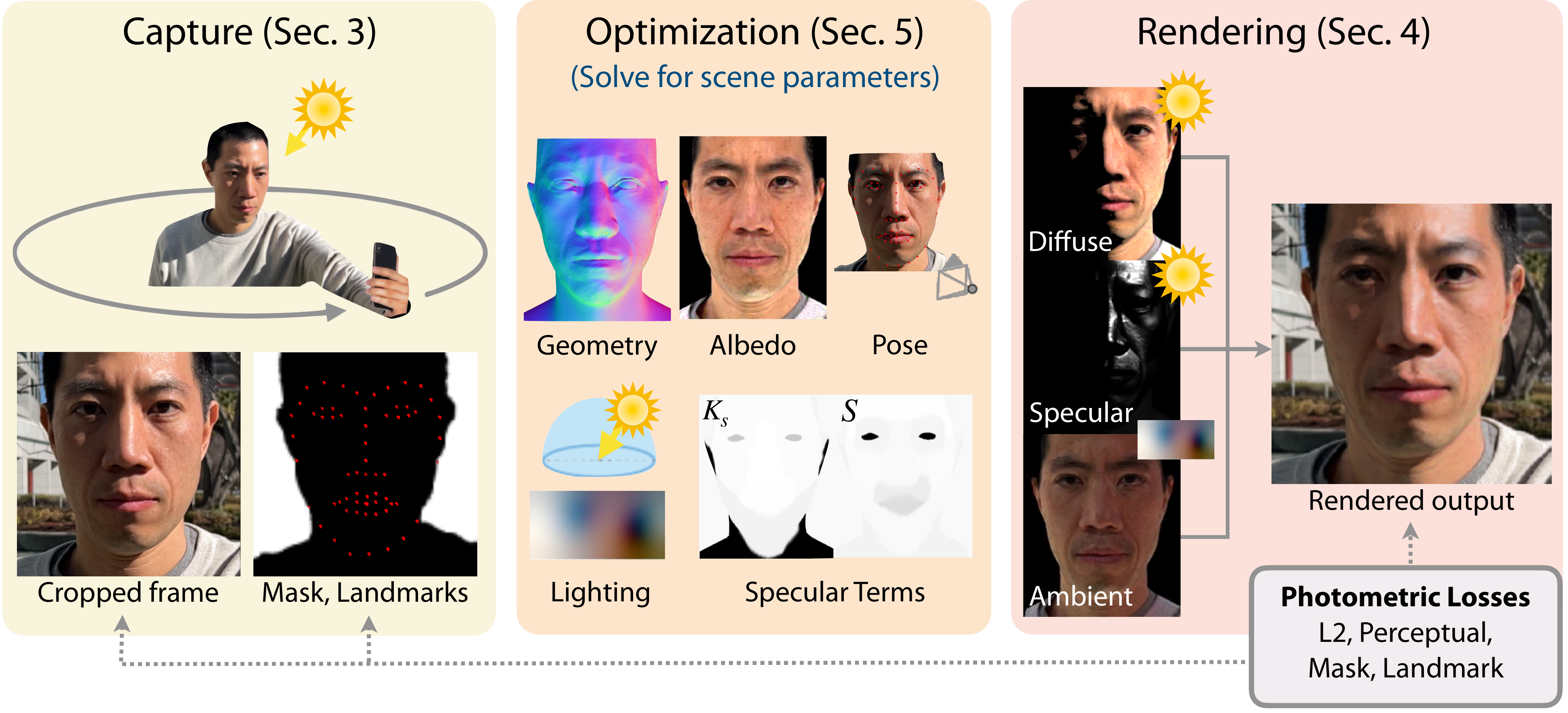}\vspace{-10pt}
\end{center}
\caption{%
\textbf{Overview.} Our method jointly reconstructs geometry, skin reflectance, lighting, and camera pose from a selfie video sequence of a person rotating under the sun. Our system begins by extracting supervisory information from the video sequence: facial landmarks, foreground alpha mattes, and camera orientations. These are used to supervise the optimization of a collection of scene parameters (full list in Sec.~\ref{subsec:photometric}) used in a physically-based renderer. The rendered output is an image consisting of diffuse, specular, and ambient light contributions. After optimization, the solved scene parameters can be used for a number of editing applications, shown in Sec.~\ref{sec:application}.
}
\vspace{-2mm}
\label{fig:overview}
\end{figure*}
\section{Overview}
\label{sec:overview}

Our method targets accurate reconstruction of scene lighting, subject geometry, and material properties from a handheld video sequence of a person rotating in place under the sun. Given a selfie video, we take a test-time optimization approach that uses the information from all frames of the video to solve for a physical model of the scene: the geometry and material properties of the face, scene lighting, and camera parameters (Fig.~\ref{fig:overview}). This physical model consists of a base face shape parameterized by a low-dimensional deformable model $X^b$, a displacement map $\Delta X$, a reflectance model with diffuse $R^d$ and specular components $R^s$, scene lighting $L_i$, and a perspective camera $C$. 
These components are explained in detail in Sec.~\ref{sec:formulation}.

After this model has been recovered, we can modify the scene and the subject parameters to re-render images.
We show several editing applications in Sec.~\ref{sec:application}: editing skin reflectance, relighting with arbitrary environment map, improving harsh lighting conditions
(by softening shadows and adding fill lights), and adjusting camera parameters to change viewpoint or manipulate perspective effects.

\section{Formulation}
\label{sec:formulation}

Given an input video, our system reconstructs the parameters of a physical model: i.e., geometry and reflectance of the subject, the scene lighting parameters, and the camera parameters.
In this section, we detail all of these parameters and describe the rendering process that turns these parameters into an image.

\paragraph{Geometry}
We denote $X_j$ as the full mesh of the subject for frame $j$, composed of a per-frame coarse mesh $X_j^b(\beta, \theta_j, \psi_j)$ and a global displacement map $\Delta {X}$. The coarse mesh $X_j^b$ is a FLAME deformable face model~\cite{li2017learning} defined by global shape code $\beta$, per-frame pose code $\theta_j$, and per-frame expression code $\psi_j$. $X_j^b$ also contains per-vertex UV coordinates, which maintain correspondence across variations in $\theta_j$ and $\psi_j$. As such, we model all our global (per-subject) spatially varying parameters in UV space, and sample values per-fragment when rasterizing.  

The displacement map $\Delta {X}$ is used to model fine details like wrinkles that cannot be represented by $X_j^b$. We displace the coarse geometry by $\Delta {X}$ at rasterization time, by sampling a displacement value per-fragment and displacing each fragment along the surface normal $N_j$ of the coarse mesh $X^b_j$. After displacement, the updated fragment positions are used to compute a new surface normal $N_j'$.

$X_j^b$ is optimized per-frame, since it accounts for subtle (and unavoidable) variations in expression and head pose during the capture, which are modeled by $\theta_j$ and $\psi_j$.
$\Delta {X}$, on the other hand, is optimized in UV-space (i.e., globally per-subject), since the deformations it encodes are invariant to the changes in expression or pose. Formally, the final geometry $X_j$ is given by:
\begin{equation}
    X_j = X^b_j(\beta, \theta_j, \psi_j) + \Delta {X} \odot N_j
\end{equation}
where $\odot$ is the Hadamard product.

\paragraph{Reflectance}
We model the skin reflectance, denoted as $R(x, \omega_i,\omega_o)\in\mathbb{R}^3$, where $x$ is a 3D point on the face geometry $X$, $\omega_i$ is the incoming light direction, and $\omega_o$ is the outgoing direction, using a diffuse and a specular component: $R=R^d+R^s$. 

The diffuse component $R^d(x, \omega_i)\in\mathbb{R}^3$ is a Lambertian reflectance model consisting of an albedo map, $a$, which we optimize as a per-subject UV-space image. For the skin's specular component, we use the Blinn-Phong model~\cite{blinn1977models}.  %add brief discussions on microfacet scattering results}:

\begin{equation}
    R^s(x, \omega_i, \omega_o) = k_s \frac{s + 2}{2 \pi} \left( h(\omega_i, \omega_o) \cdot n(x) \right)^s \label{eqn:brdf}\,
\end{equation}
where $h(\omega_i, \omega_o) = \text{normalize}(\omega_i + \omega_o)$ is the half vector, $k_s$ is the specular intensity, $s$ is the specular exponent, and $(s+2)/(2\pi)$ is the normalization term for the reflection lobe to integrate to $1$.
% We optimize  as a per-subject UV-space parameter map
Following~\cite{weyrich2006analysis}, we segment the UV-space map into 10 segmented specular reflectance clusters. We then optimize for a spatially-varying pair of values $(s,k_s)$ per-cluster, enabling varying shininess across the face.

While Blinn-Phong does not model many complex effects such as subsurface scattering, our experiments with other models for facial reflectance, such as microfacet models \cite{walter2007microfacet}, show no significant quality improvements, and often introduce unstable training. More analysis is provided in the supplementary material.

\paragraph{Lighting}
We use a sun-sky model to represent lighting as the sum of an ``ambient'' environment map and the sun: $L_i(x, \omega_i)=L_i^\text{amb}(\omega_i)+L_i^\text{sun}(\omega_i)$.
Note neither $L_i^\text{amb}(\omega_i)$ nor $L_i^\text{sun}(\omega_i)$ depends on the 3D point $x$, since we model both as directional lights.
Optimization-wise, our lighting parameters consist of a $16\times32\times3$ environment map for ambient lighting, the sun direction $p^\text{sun} \in S^3$, and the scalar sun intensity $k^\text{sun}$.
We fix the sun color to white $[1, 1, 1]$ in our lighting model to resolve the albedo-illumination ambiguity.

\subsection{Rendering}

We calculate the outgoing radiance $L_o$ at 3D location $x$ as viewed from viewing direction $\omega_o$ as:

\begin{align}
    &L_o(x, \omega_o)  \nonumber\\
    &= \int_\Omega V(x, \omega_i) L_i(x, \omega_i) \odot R(x, \omega_i, \omega_o) \left( \omega_i \cdot n(x) \right) d\omega_i\\
    &= \sum_{\omega_i} V(x, \omega_i) \Big(L_i^\text{amb}(\omega_i) \odot R^d(x, \omega_i) \\ 
    &+ L_i^\text{sun}(\omega_i) \odot R^d(x,\omega_i) + \nonumber
    L_i^\text{amb}(\omega_i) \odot R^s(x, \omega_i, \omega_o) \\ 
    &+ L_i^\text{sun}(\omega_i) \odot R^s(x, \omega_i, \omega_o) \Big) \left( \omega_i \cdot n(x) \right) \triangle \omega_i \label{eqn:render}\,
\end{align}
\normalsize
where $V(x,\omega_i)$ is the light visibility at $x$ from $\omega_i$, and $L_i(x,\omega_i)$ is the incoming radiance reaching $x$ from $\omega_i$.
We ignore the specular reflection caused by the ambient lighting, \ie, $L_i^\text{amb}(\omega_i) \odot R^s(x, \omega_i, \omega_o)$, since it is much weaker than the specular reflection of the sun.
In the next subsections, we will group the terms into a diffuse contribution $L^d_o$ and a specular contribution $L^s_o$: $L_o=L^d_o+L^s_o$.
For the final rendered color value, we apply the Reinhard operator~\cite{reinhard2002parameter} and a gamma correction of $\gamma=2.2$ to $L_o$ to convert from linear to sRGB space.

\paragraph{Diffuse contribution}
The diffuse contribution $L^d_o$ is then given by only the diffuse terms of \eqn{eqn:render}:
\begin{align}
    L_o^d(x) &= 
    \sum_{\omega_i} L_i^\text{amb}(\omega_i) \odot \frac{a(x)}{\pi}(\omega_i \cdot n(x)) \Delta\omega_i \nonumber\\
    &+ V(x, p^\text{sun}) k^\text{sun}[1, 1, 1] \odot \frac{a(x)}{\pi} (p^\text{sun} \cdot n(x)) \Delta p^\text{sun}\,
    \label{eq:dif_c5}
\end{align}
where $a(x)$ is the albedo at point $x$, $k^\text{sun}$ is the (optimized) sun intensity, and $p^\text{sun}$ is the (optimized) sun direction. The sun is modeled as a directional light source, so the second summation can be simplified to a single term, only in the direction of $p^\text{sun}$.  We additionally optimize for a high-dynamic-range (HDR) environment map $E \in \mathbb{R}^{16 \times 32 \times 3}$, from which values of $L_i^\text{amb}$ are sampled.

\paragraph{Specular contribution}
The specular contribution $L^s_o$ at each pixel is given by only the specular term due to the sun in \eqn{eqn:render} (recall that we ignore the specular ambient term due to its weak contribution):
\small
\begin{align}
    L^s_o(x, \omega_o) &= 
    V(x, p^\text{sun}) k^\text{sun} [1, 1, 1] k_s \frac{s + 2}{2 \pi} \nonumber\\
    &\left(h(p^\text{sun}, \omega_o) \cdot n(x) \right)^s \left( p^\text{sun} \cdot n(x) \right) \Delta p^\text{sun} \,
    \label{eq:sp_c6}
\end{align}
\normalsize
where we have substituted \eqn{eqn:brdf} and reduced the summation to just one term at $p^\text{sun}$ (since $L_i^\text{sun}$ is $0$ elsewhere). %\todo{We also tried a microfacet reflectance model~\cite{walter2007microfacet}. It has comparable visual quality to the bling phong BRDF. We decided to adopt the bling phong BRDF due to its optimization simplicity and stability.}

\paragraph{Shadow map}
In order to generate a map of self-occluded shadows, we perform two passes of rasterization: first, we render a $z$-buffer from a virtual orthographic camera aligned with the sun direction, $p^\text{sun}$, and then, when rasterizing a given camera viewpoint, compare all fragment positions  $d_\text{hit}$ to the light's $z$-buffer $d_\text{shadow}$.
To avoid precision issues and ensure smooth gradients for back-propagation, we implement a soft comparison as follows in generating shadow/visibility maps:
\begin{equation}
    V(x, \omega_i) = 1 - \text{sigmoid}\left( k (d_\text{hit} - d_\text{shadow} \times b) \right) \,
    \label{eq:soft}
\end{equation}
where $k$ is the falloff slope, and $b$ the tolerance. We use $k=800$ and $b=1.0015$.

\section{Optimization}
\label{sec:optimization}

% \TODO{initialization: camera conversion, DECA geometry, reflectance}

% \TODO{training scheme: the order of optimization, parameters, loss functions for each stage }

The described physical model contains a large number of parameters to be optimized, controlling scene elements like lighting, geometry, pose, and texture.
Unfortunately, na\"ively optimizing all these parameters from scratch does not result in an optimal solution, since the final observed appearance of the face can often be explained variously through changes to geometry, material properties, lighting or camera parameters, making optimization severely under-constrained and ambiguous.
Therefore, we adopt a two-stage optimization approach, through which parameters are gradually enabled.
In this section, we describe this process and the relevant losses that guide optimization.

\subsection{Coarse alignment}
\label{sec:optimization-stage1}
Our system begins by using an off-the-shelf network (DECA~\cite{feng2021learning}) to generate, for each input image, a set of shape parameters $\beta$, pose parameters $\theta_j$, and expression parameters $\psi_j$ of a FLAME face model~\cite{li2017learning}, as well as the relative pose parameters of the virtual camera observing the 3D face. Unfortunately, as with many other single-image facial geometry estimators, DECA assumes an orthographic projection model and therefore cannot accurately recover geometry for our selfie capture sequences, which contain heavy perspective effects (Figure~\ref{fig:ablation-2-stage}).
Without a good initialization for geometry, optimization of lighting and material properties seldom converges to an optimal solution due to the heavily ambiguous nature of our optimization problem.

To circumvent this issue, we employ a first stage of optimization where we only optimize for the parameters of a perspective camera (with a known focal length, extracted from input metadata) and the face geometry parameters $(\beta, \theta_j, \psi_j)$.
As initialization for this optimization process, we use the predicted DECA values for each frame's pose $\theta_j$ and expression $\psi_j$, but set all frames to the average predicted shape $\beta_\text{avg} = \frac{1}{N}\sum_j \beta_j$, since the identity remains constant across all frames.
To convert DECA's orthographic camera to a perspective camera, we additionally optimize for an unknown scale $S$ and translation $T_j$, which are initialized to empirically chosen values $S=2.6e4$, $T_j=(0,0,1.5e5)$.
During optimization, the face shape $\beta$ and scale $S$ are shared across all frames, while camera pose $T_j$, expression $\psi_j$, and pose $\theta_j$ are optimized per-frame.
Note that DECA controls the relative orientation of the camera and the face by varying the pose code $\theta$ instead of the camera rotation.
We adopt this formulation and keep the camera orientation fixed relative to the face.
The global orientation of the camera at each frame (and therefore the face) is extracted from the capture video, either through a structure-from-motion system or IMU measurements commonly available on a smartphone.

We use two losses to guide this optimization: a mask loss $L_\text{mask}$ and a landmark loss $L_\text{lmk}$.
The FLAME model includes 3D facial landmark points, corresponding to the standard 68-point facial landmarks set~\cite{sagonas2013300} used in facial tracking. Our landmark loss minimizes the L1 distance between the 2D projection of these 3D landmarks (into the input camera viewpoint) and 2D landmarks estimated from the input frame by a 2D landmark detector HRNets~\cite{wang2020deep}.

The facial landmarks provide a stong constraint on facial feature alignment, but are sparse, and therefore cannot constrain the overall shape or boundary of the mesh. 
To supplement it, we include a silhouette loss $L_{\text{mask}}$, which penalizes the L2 difference between the rasterized mask of the mesh $I_\text{sil}$ and the semantic segmentation mask $I_\text{mask}$ of the input image, using an off-the-shelf semantic segmentation network~\cite{lin2022robust} trained to segment humans in portrait photographs.

The final pose loss is then: $L_\text{pose} = L_\text{mask} +  L_\text{lmk}$, optimized using an ADAM optimizer \cite{kingma2014adam}. See supplemental for optimization parameters. 
\begin{figure*}[t!]
\begin{center}

%\begin{subfigure}[b]{0.12\linewidth}
%\includegraphics[width=\linewidth]{fig/comparisons/dan_rl_nearest.png} %
%\end{subfigure}
\begin{subfigure}[b]{0.12\linewidth}
\includegraphics[width=\linewidth]{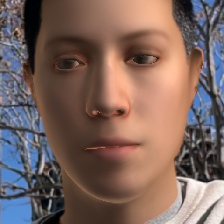} %
\end{subfigure}
\begin{subfigure}[b]{0.12\linewidth}
\includegraphics[width=\linewidth]{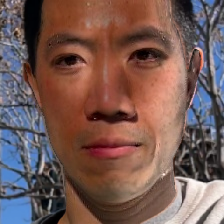} %
\end{subfigure}
\begin{subfigure}[b]{0.12\linewidth}
\includegraphics[width=\linewidth]{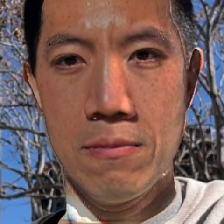} %
\end{subfigure}
\begin{subfigure}[b]{0.12\linewidth}
\includegraphics[width=\linewidth]{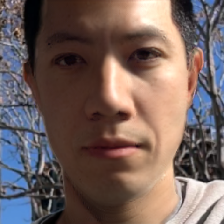} %
\end{subfigure}
\begin{subfigure}[b]{0.12\linewidth}
\includegraphics[width=\linewidth]{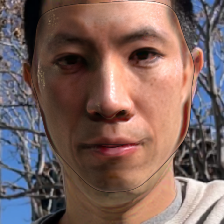} %
\end{subfigure}
\begin{subfigure}[b]{0.12\linewidth}
\includegraphics[width=\linewidth]{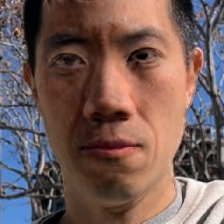} %
\end{subfigure}
\begin{subfigure}[b]{0.12\linewidth}
\includegraphics[width=\linewidth]{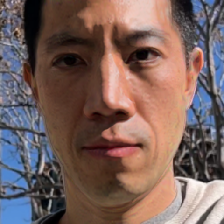} %
\end{subfigure}
\begin{subfigure}[b]{0.12\linewidth}
\includegraphics[width=\linewidth]{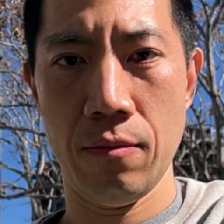} %
\end{subfigure}\\
%\begin{subfigure}[b]{0.12\linewidth}
%\includegraphics[width=\linewidth]{fig/comparisons/florian_rl_nearest.png} %
%\caption{Nearest}
%\end{subfigure}
\begin{subfigure}[b]{0.12\linewidth}
\includegraphics[width=\linewidth]{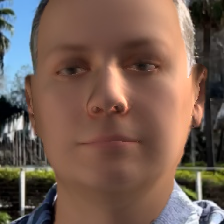} %
\caption{\scalebox{0.92}[1.0]{DECA}~\cite{feng2021learning}}
\end{subfigure}
\begin{subfigure}[b]{0.12\linewidth}
\includegraphics[width=\linewidth]{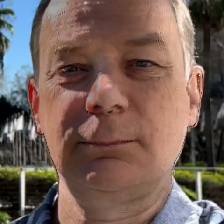} %
\caption{DPR~\cite{zhou2019deep}}
\end{subfigure}
\begin{subfigure}[b]{0.12\linewidth}
\includegraphics[width=\linewidth]{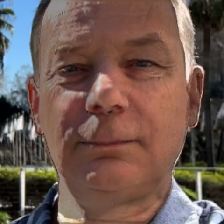} %
\caption{GCFR~\cite{hou2022face}}
\end{subfigure}
\begin{subfigure}[b]{0.12\linewidth}
\includegraphics[width=\linewidth]{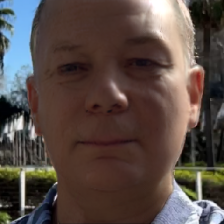} %
\caption{TR~\cite{pandey2021total}}
\end{subfigure}
\begin{subfigure}[b]{0.12\linewidth}
\includegraphics[width=\linewidth]{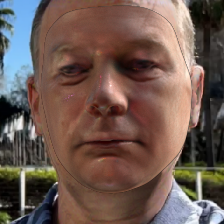} %
\caption{NextFace~\cite{dib2021towards}}
\end{subfigure}
\begin{subfigure}[b]{0.12\linewidth}
\includegraphics[width=\linewidth]{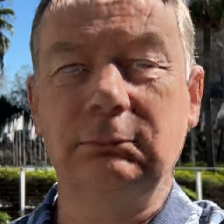} %
\caption{NLT~\cite{zhang2021neural}}
\end{subfigure}
\begin{subfigure}[b]{0.12\linewidth}
\includegraphics[width=\linewidth]{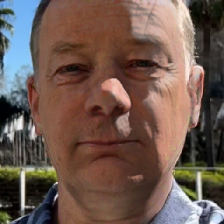} %
\caption{Ours}
\end{subfigure}
\begin{subfigure}[b]{0.12\linewidth}
\includegraphics[width=\linewidth]{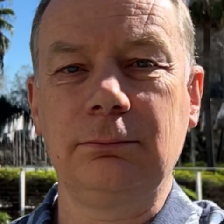} %
\caption{GT}
\end{subfigure}
\\
\caption{\textbf{Qualitative: Relighting}. A comparison of our method at rendering a new (unseen) lighting environment (h). Our method is able to realistically synthesize the novel lighting condition, including cast shadows and specularity, and nearly matches the (unseen) target reference image. See supplement for additional details on experimental setup and analysis of results.}
%\vspace{-4em}
\label{fig:qualitative-relighting}
\end{center}

\end{figure*}

\begin{figure*}
\begin{center}

\begin{subfigure}[b]{0.16\linewidth}
\includegraphics[width=\linewidth]{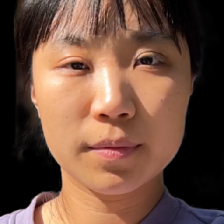} %
\end{subfigure}
\begin{subfigure}[b]{0.16\linewidth}
\includegraphics[width=\linewidth]{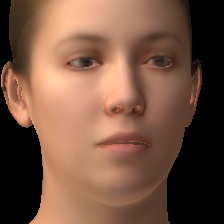} %
\end{subfigure}
\begin{subfigure}[b]{0.16\linewidth}
\includegraphics[width=\linewidth]{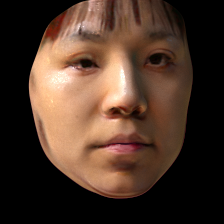} %
\end{subfigure}
\begin{subfigure}[b]{0.16\linewidth}
\includegraphics[width=\linewidth]{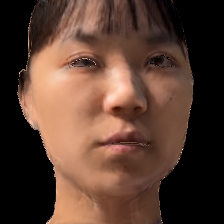} %
\end{subfigure}
\begin{subfigure}[b]{0.16\linewidth}
\includegraphics[width=\linewidth]{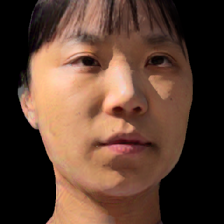} %
\end{subfigure}
\begin{subfigure}[b]{0.16\linewidth}
\includegraphics[width=\linewidth]{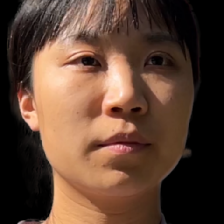} %
\end{subfigure}\\

\begin{subfigure}[b]{0.16\linewidth}
\includegraphics[width=\linewidth]{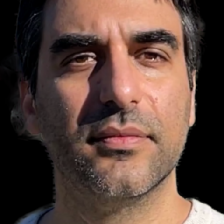} %
\caption{Nearest view}
\end{subfigure}
\begin{subfigure}[b]{0.16\linewidth}
\includegraphics[width=\linewidth]{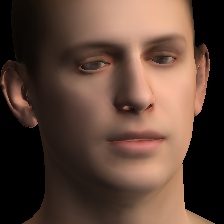} %
\caption{DECA \cite{feng2021learning}}
\end{subfigure}
\begin{subfigure}[b]{0.16\linewidth}
\includegraphics[width=\linewidth]{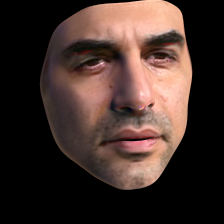} %
\caption{NextFace\cite{dib2021towards}}
\end{subfigure}
\begin{subfigure}[b]{0.16\linewidth}
\includegraphics[width=\linewidth]{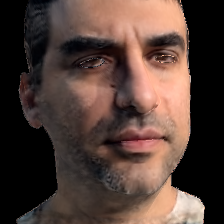} %
\caption{NLT \cite{zhang2021neural}}
\end{subfigure}
\begin{subfigure}[b]{0.16\linewidth}
\includegraphics[width=\linewidth]{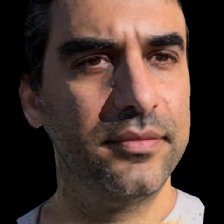} %
\caption{Ours}
\end{subfigure}
\begin{subfigure}[b]{0.16\linewidth}
\includegraphics[width=\linewidth]{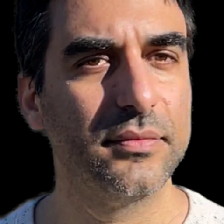} %
\caption{GT}
\end{subfigure}\\

\caption{\textbf{Qualitative: view synthesis}. A comparison of our method at the task of generating an image from an unseen viewpoint (f), having only seen a limited collection of input viewpoints. See supplement for more details on experimental setup and analysis of results.}

\label{fig:qualitative-view}

% \vspace{-5mm}

\end{center}
\end{figure*}

\subsection{Photometric optimization}
\label{subsec:photometric}
Once the 3D model and camera parameters are approximately aligned, we proceed to the second stage of optimization, in which we optimize the precise facial geometry, lighting, and reflectance properties.
All the parameters optimized in the first stage (Section~\ref{sec:optimization-stage1}) remain as free variables.
In total, the parameters optimized during this stage include: \textbf{Lighting parameters}: (1) $p^\text{sun}$, the global sun direction, (2) $E$, the global environment map, (3) $k^\text{sun}$, the global sun intensity, \textbf{Facial geometry parameters}: (4) $\beta$, the global FLAME shape code, (5) $\psi_j$, the per-frame expression code, (6) $\theta_j$, the per-frame pose code, (7) $\Delta X$, the global deformation map, \textbf{Material properties}: (8) $k_s$, the global, spatially-varying specular intensity, (9) $s$, the global, spatially-varying specular roughness, (10) $a$, the global, spatially-varying surface albedo, \textbf{Camera pose parameters}: (11) $T_j$, the per-frame perspective camera translation, and (12) $S$, the global scene scale.

During optimization, we randomly select a frame $j$, render the face using a differentiable rasterizer \cite{ravi2020accelerating} and the equations described in Section~\ref{sec:formulation} to get the rendered image $\hat{I}$.
In addition to the previously defined landmark and mask losses, we include L2 and VGG \cite{johnson2016perceptual} photometric losses, comparing the original and reconstructed images:
\begin{equation}
    L_\text{photo} = || \hat{I}_j \cdot I_\text{sil}  - I_j \cdot I_\text{mask}||_2
\end{equation}
We also include an L2 regularization $L_E$ and L2-smoothness regularization $L_{E\text{s}}$ on the reconstructed environment map, to encourage the majority of the lighting to be explained by direct sunlight and to aid in disentanglement of the sun and ambient lighting.
The total optimized loss becomes:
\begin{align}
    L &= \lambda_\text{mask}L_\text{mask} + \lambda_\text{lmk} L_\text{lmk} + \lambda_E L_{E} + \lambda_{E\text{s}} L_{E\text{s}} \nonumber \\
    &+ \lambda_{\text{VGG}} L_{\text{VGG}} + \lambda_{\text{photo}} L_{\text{photo}}\
\end{align}
with $\lambda_{\text{mask}}, \lambda_{\text{lmk}} = 0.05, \lambda_\text{VGG} = 0.005, \lambda_E = 0.01, \lambda_{E\text{s}}, \lambda_\text{photo} = 1$.
Additional optimization details are provided in the supplemental materials.

\section{Evaluation}
\label{sec:evaluation}

In this section, we detail quantitative and qualitative experiments comparing our approach with state-of-the-art methods and ablated variants of our method. 

%\todo{We added some baseline evaluations. For novel-view synthesis, we now have NextFace~\cite{dib2021towards}. For relighting, we now have GCFR~\cite{hou2022face} and NextFace~\cite{dib2021towards}. We also compared with NVPR~\cite{zhang2021neuralvideo} qualitatively in the suppl.}

\paragraph{Baseline comparisons}
We evaluate our method on the tasks of novel-view synthesis and relighting.
For novel-view synthesis, we compare our method with DECA~\cite{feng2021learning}, Neural Light Transport (NLT)~\cite{zhang2021neural}, and NextFace~\cite{dib2021towards}.
For relighting, we compare with DECA, NLT, NextFace, GCFR~\cite{hou2022face}, image-based methods Deep Single Image Portrait Relighting (DPR) \cite{zhou2019deep}, Total Relighting (TR)~\cite{pandey2021total} and NVPR~\cite{zhang2021neuralvideo}.
Additional comparisons and details on the experimental setups are provided in the supplemental materials.

\begin{figure*}
\begin{center}

\includegraphics[width=1.00\linewidth]{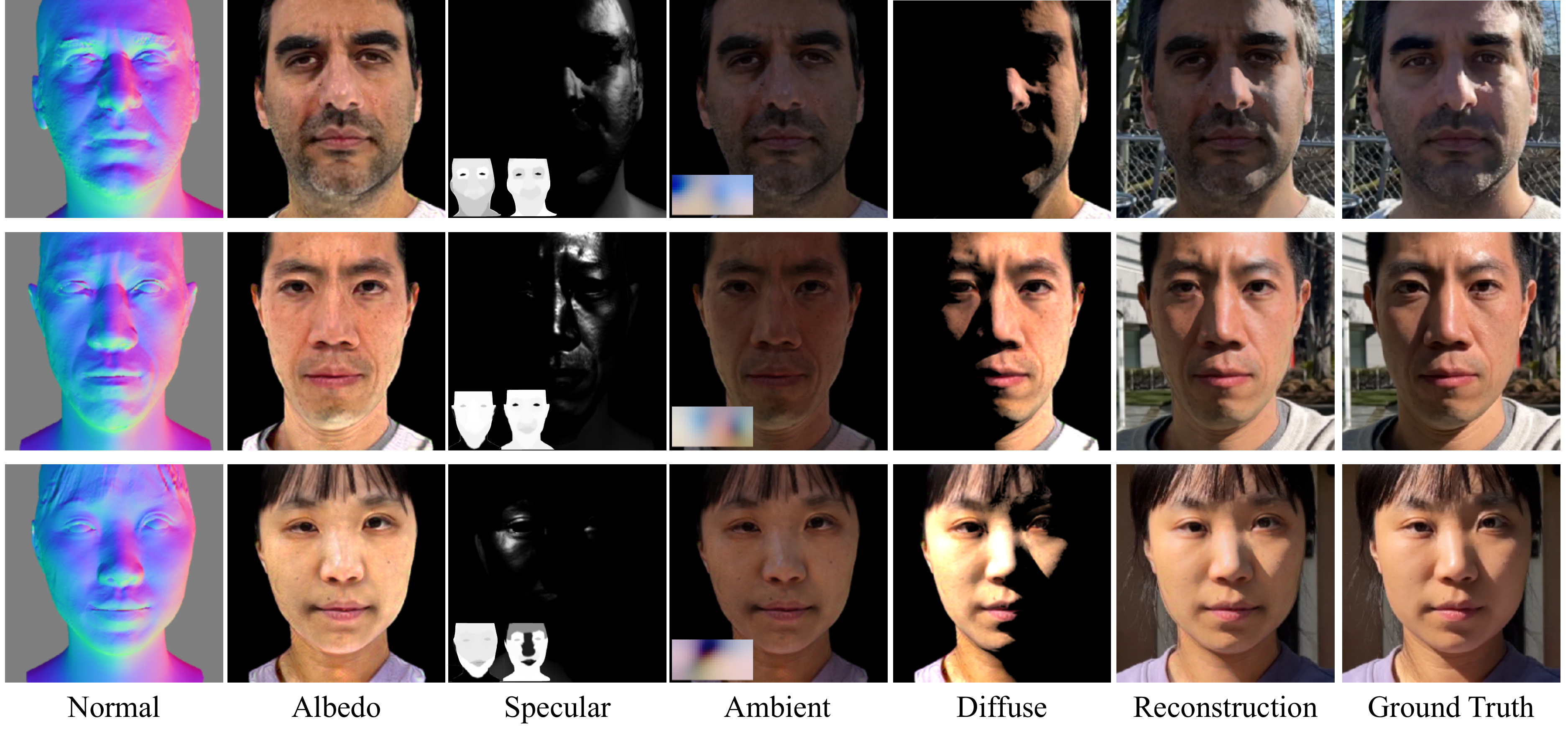}

\caption{\textbf{Decomposition.} We show all the components which comprise our final rendered image to demonstrate that our method not only closely recreates the ground truth image (reproducing realistic highlights and shadows), but also performs a meaningful decomposition of lighting components and facial geometry. Note that our reconstructed surface normals include high frequency details specific to each subject, like wrinkles and birthmarks, which are used in computation of the shadows and specular reflections.}
\vspace{-7mm}

\label{fig:qualitative-recon}
\end{center}
\end{figure*}

We present qualitative comparisons for relighting in Figure~\ref{fig:qualitative-relighting} and novel view synthesis in Figure~\ref{fig:qualitative-view}.
Quantitative comparisons on these images are provided in Table~\ref{tbl:quant}.
These testing images consist of (1) a multi-view capture of the face, in which the subject remains still and the camera is moved to novel viewpoints in the same environment as the original capture, and (2) front-facing sequences in novel environment lighting and unseen sun positions.
All testing images are not seen during training of our method, NLT or NextFace.
The results shown in Figures~\ref{fig:qualitative-relighting}~and~\ref{fig:qualitative-view} as well as Table~\ref{tbl:quant} clearly demonstrate that our method outperforms all the baselines at both relighting and view synthesis.
Single-image methods (DECA, GCFR, DPR, TR) can generalize to other subjects, but fail to recover more faithful and physically accurate facial details.
Comparison with multi-image methods (NLT, NextFace) demonstrates that SunStage is a better reconstruction system.
Additional analysis of the comparisons is provided in the supplemental materials.

\begin{table}[t!]
\footnotesize
    \setlength{\tabcolsep}{0.15\tabcolsep}% Shrink \tabcolsep by 30%
    \centering
    
    \resizebox{\linewidth}{!} {%
    \begin{tabular}{@{} l|ccc||ccc}
     &  \multicolumn{3}{c}{Relighting}  ~& \multicolumn{3}{c}{Novel view synthesis}   \\ \hline
   ~ &~ \!PSNR$\uparrow$\! ~&~ \!SSIM$\uparrow$\! ~&~ \!LPIPS$\downarrow$\!   ~& ~\!PSNR$\uparrow$\!~ & ~\!SSIM$\uparrow$\! ~&~ \!LPIPS$\downarrow$\!  \\ \hline
    DECA \cite{feng2021learning} & 16.41 & 0.69 & 0.25 & 16.64 & 0.66 & 0.29\\
    GCFR \cite{hou2022face} & 16.97 & 0.70 & 0.20 & - & - & -\\
    DPR \cite{zhou2019deep} & 19.03 & 0.72 & 0.19 & - & - & -\\
    NLT \cite{zhang2021neural} & 20.15 & 0.75 & 0.18 & \cellcolor{JonYellow}22.27 & \cellcolor{JonOrange}0.79 & 0.15\\
    Total Relighting~\cite{pandey2021total} & 20.24 & \cellcolor{JonOrange} 0.79 & 0.16 & - & - & - \\
    NextFace \cite{dib2021towards} & \cellcolor{JonOrange}22.98 & 0.76 & 0.15 & \cellcolor{JonOrange}22.55 & 0.75 & 0.15\\
    
    \hline
    
    Ours & \cellcolor{JonRed}23.64 & \cellcolor{JonRed}0.83 & \cellcolor{JonRed}0.10 & \cellcolor{JonRed}25.28 & \cellcolor{JonRed}0.84 & \cellcolor{JonRed}0.09 \\

    \hline 
    
    Ours w/o coarse & 17.83 & 0.66 & 0.23 & 19.65 & 0.70 & 0.17\\
    Ours w/o SV $k_s, s$ & 21.31 & 0.77 & \cellcolor{JonYellow}0.13 & 21.94 & 0.77 & \cellcolor{JonYellow}0.12\\
    Ours w/o $L_\text{mask}$, $L_\text{lmk}$ & 16.46 & 0.61 & 0.30 & 18.83 & 0.68 & 0.20\\
    Ours w/o $L_\text{mask}$ & 20.13 & 0.75 & 0.15 & 20.54 & 0.74 & 0.15\\
    Ours w/o opt. $(\beta, \theta_i, \phi_i)$ \ & 18.67 & 0.69 & 0.19 & 18.28 & 0.66 & 0.19\\
    Ours w/o soft shadow & \cellcolor{JonYellow}21.46 & \cellcolor{JonYellow}0.77 & \cellcolor{JonOrange}0.13 & 22.05 & \cellcolor{JonYellow}0.77 & \cellcolor{JonOrange}0.12\\
    Ours w/o $\Delta X$ & 21.16 & 0.75 & 0.15 & 21.80 & 0.75 & 0.14\\
    \end{tabular}%
    }
    \caption{\textbf{Quantitative comparison}. Comparison of our method on the tasks of novel view synthesis and relighting. See Section~\ref{sec:evaluation} for a description of the ablated variants.}
    \label{tbl:quant}

\end{table}

\paragraph{Disentanglement}
In Figure~\ref{fig:qualitative-recon}, we demonstrate how SunStage decomposes the appearance of a portrait photograph into different components: specular, diffuse, and ambient.
We also visualize the surface normal, albedo, and other intermediate representations to show that our method is able to effectively recover a physically plausible reconstruction of the real world and disentangle the different components that contribute to the final appearance. We further validate the quality of the reconstructed geometry and materials in the supplementary material.

\begin{figure}
\begin{center}
\begin{subfigure}[t]{0.24\linewidth}
\includegraphics[width=\linewidth]{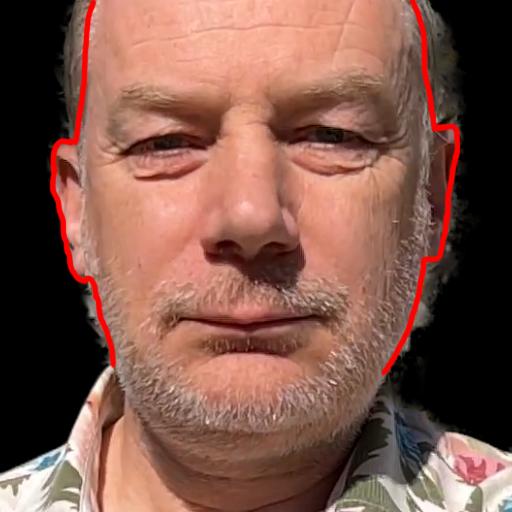} %
\caption{Input image}
\end{subfigure}
\begin{subfigure}[t]{0.24\linewidth}
\includegraphics[width=\linewidth]{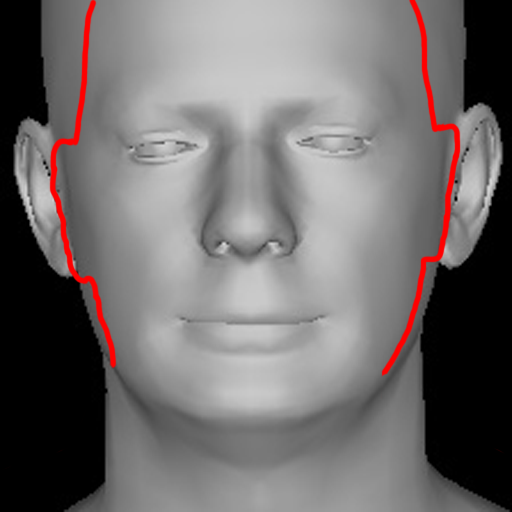}
\caption{DECA \cite{feng2021learning}}
\end{subfigure}
\begin{subfigure}[t]{0.24\linewidth}
\includegraphics[width=\linewidth]{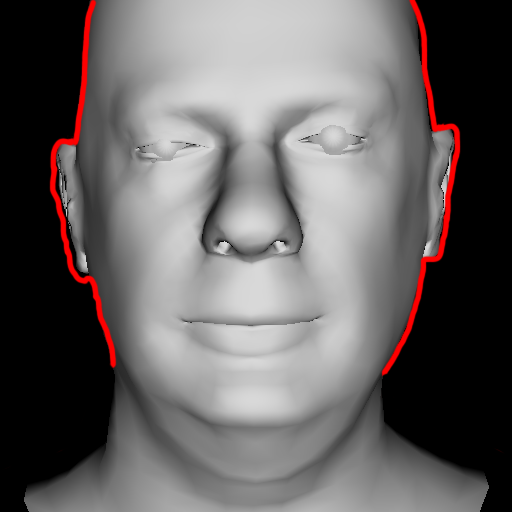} %
\caption{Stage 1}
\end{subfigure}
\begin{subfigure}[t]{0.24\linewidth}
\includegraphics[width=\linewidth]{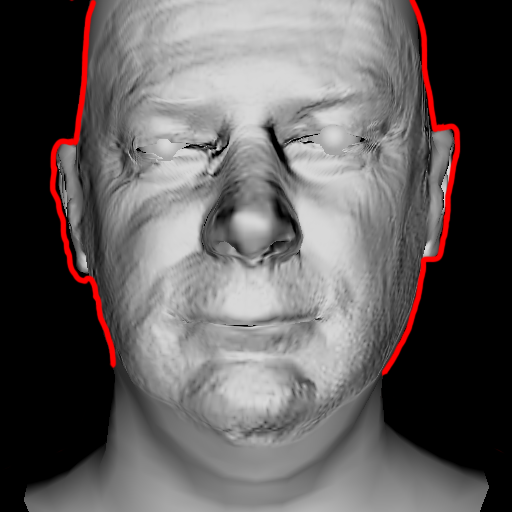} %
\caption{Final}
\end{subfigure}
\caption{\textbf{Perspective}. DECA's assumption of an orthographic camera is broken by the strong perspective effects in selfies, causing poor alignment (b) with input images (a). Our first stage of optimization (c) (Sec.~\ref{sec:optimization-stage1}) improves alignment by solving for the parameters of a perspective camera and refined shape parameters. In the second stage we additionally optimize for a displacement map $\Delta X$ to produce our final shape with finer geometric details like wrinkles (d). Red line added to highlight alignment with (a).}
\label{fig:ablation-2-stage}

%\vspace{-7mm}

\end{center}
\end{figure}
\paragraph{Ablation studies}
In addition to our comparisons with the state-of-the-art baselines, we also compare with ablated variants of our own method.
In particular, we include seven such experiments in Table~\ref{tbl:quant}: our method (1) without the initial first stage of coarse geometric alignment, i.e., directly optimizing both geometric and photometric parameters from the start, (2) without the spatially varying specular parameters, instead using a single global scalar $s$ and $k_s$, (3) without the geometric alignment losses $L_\text{mask}$ and $L_\text{lmk}$, (4) without just $L_\text{mask}$, (5) without shape optimization, i.e., keeping the initial shape code predicted by DECA, (6) without soft shadow computation, i.e., using a hard z-buffer comparison to compute a shadow map instead of our soft comparison operator in Equation~\ref{eq:soft}, and (7) without the displacement map $\Delta X$.
Visual results for each of these variants are provided in the supplemental material.

\section{Applications}
\label{sec:application}

\begin{figure}
\begin{center}

\begin{subfigure}[t]{0.24\linewidth}
\includegraphics[width=\linewidth]{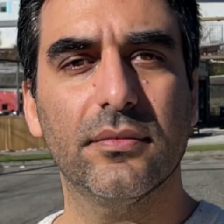} %
\end{subfigure}
\begin{subfigure}[t]{0.24\linewidth}
\includegraphics[width=\linewidth]{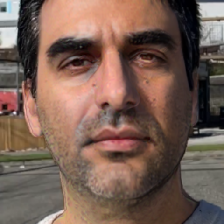} %
\end{subfigure}
\begin{subfigure}[t]{0.24\linewidth}
\includegraphics[width=\linewidth]{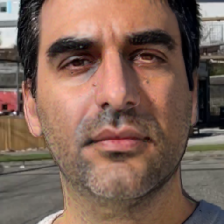} %
\end{subfigure}
\begin{subfigure}[t]{0.24\linewidth}
\includegraphics[width=\linewidth]{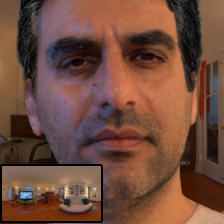} %
\end{subfigure}\\

\begin{subfigure}[t]{0.24\linewidth}
\includegraphics[width=\linewidth]{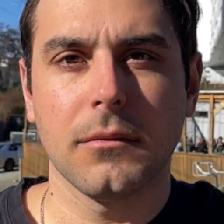} %
\caption{Input}
\end{subfigure}
\begin{subfigure}[t]{0.24\linewidth}
\includegraphics[width=\linewidth]{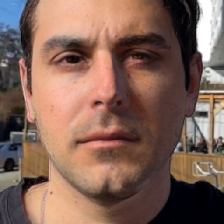} %
\caption{Soft shadows}
\end{subfigure}
\begin{subfigure}[t]{0.24\linewidth}
\includegraphics[width=\linewidth]{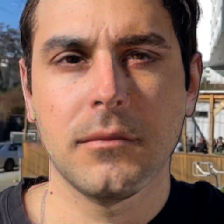} %
\caption{{(b) + fill light}}
\end{subfigure}
\begin{subfigure}[t]{0.24\linewidth}
\includegraphics[width=\linewidth]{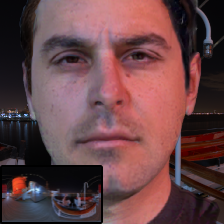} %
\caption{Swap scene}
\end{subfigure}\\

\caption{\textbf{Adjusting lighting parameters}. We can adjust the recovered scene parameters to improve the lighting conditions in an input image (a) by softening the harsh shadows cast by the nose (b), adding a fill light to brighten the shaded region (b), or replacing the environment altogether (d).}
\label{fig:relight}

\vspace{-10pt}

\end{center}
\end{figure}

\paragraph{Relighting}
We demonstrate two types of relighting applications: (1) lighting modification and (2) lighting replacement. Practical lighting modification is common in portrait photography when the lighting conditions are not ideal, e.g., when direct sunlight casts undesirable harsh shadows with high contrast. A common practice is to make the light source larger and more diffuse by using a scrim or bounce card. In Fig.~\ref{fig:relight}b, we show that by virtually increasing the size and spreading the energy of our reconstructed lighting source (\ie, the sun), we are able to \emph{soften} the shadows and re-render a more visually pleasing face. %gg and therefore a more appealing look. 
Another approach to reducing the effects of harsh shadows is adding local fill lights, which reduces the contrast between the lit and shaded regions (Fig.~\ref{fig:relight}c). Alternatively, fill lights can also be used for artistic purposes, to create dramatic lighting effects (Fig.~\ref{fig:teaser}b). Finally, replacing the scene lighting with that of a novel environment~(Fig.~\ref{fig:relight}d) is a necessary step in realistically inserting a captured subject into a virtual scene, which is useful for visual effects and VR applications.

\begin{figure}
\begin{center}

\begin{subfigure}[t]{0.24\linewidth}
\includegraphics[width=\linewidth]{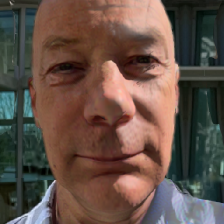} %
\end{subfigure}
\begin{subfigure}[t]{0.24\linewidth}
\includegraphics[width=\linewidth]{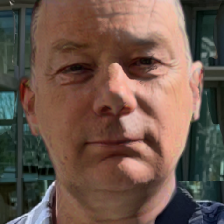} %
\end{subfigure}
\begin{subfigure}[t]{0.24\linewidth}
\includegraphics[width=\linewidth]{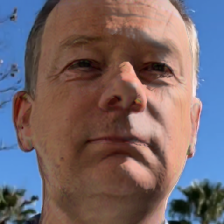} %
\end{subfigure}
\begin{subfigure}[t]{0.24\linewidth}
\includegraphics[width=\linewidth]{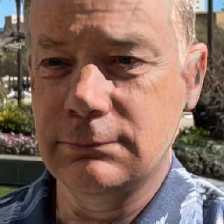} %
\end{subfigure}\\

\begin{subfigure}[t]{0.24\linewidth}
\includegraphics[width=\linewidth]{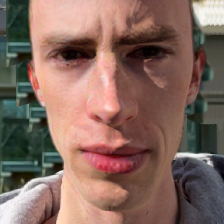} %
\caption{Short focal}
\end{subfigure}
\begin{subfigure}[t]{0.24\linewidth}
\includegraphics[width=\linewidth]{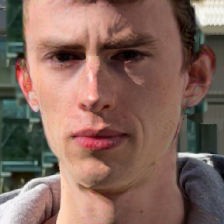} %
\caption{Long focal}
\end{subfigure}
\begin{subfigure}[t]{0.24\linewidth}
\includegraphics[width=\linewidth]{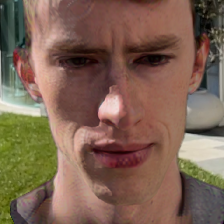} %
\caption{Novel view 1}
\end{subfigure}
\begin{subfigure}[t]{0.24\linewidth}
\includegraphics[width=\linewidth]{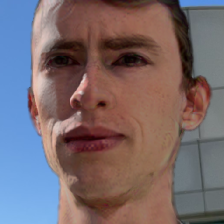} %
\caption{Novel view 2}
\end{subfigure}\\

\caption{\textbf{Changing camera parameters}. We can change the recovered camera parameters to render novel views (c,d) or change the focal length (a,b).}
\label{fig:app-camera-view}

\vspace{-15pt}

\end{center}
\end{figure}

\paragraph{View Synthesis}
In Figure~\ref{fig:app-camera-view}, we show that our reconstructed 3D model of the face can be used to synthesize new views by manipulating the viewpoint of the camera. We can also change other camera parameters, such as the focal length, to reduce the perspective effects on the face, which is often desirable for selfie images that contain significant facial distortion due to perspective.

\begin{figure}
\begin{center}

\begin{subfigure}[t]{0.24\linewidth}
\includegraphics[width=\linewidth]{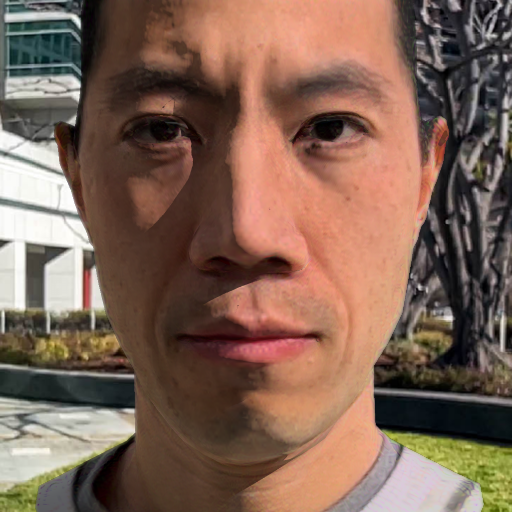} %
\caption{Less shinier}
\end{subfigure}
\begin{subfigure}[t]{0.24\linewidth}
\includegraphics[width=\linewidth]{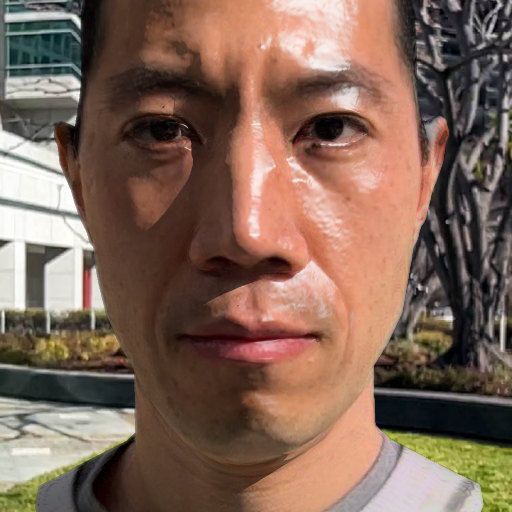}
\caption{More shinier}
\end{subfigure}
\begin{subfigure}[t]{0.24\linewidth}
\includegraphics[width=\linewidth]{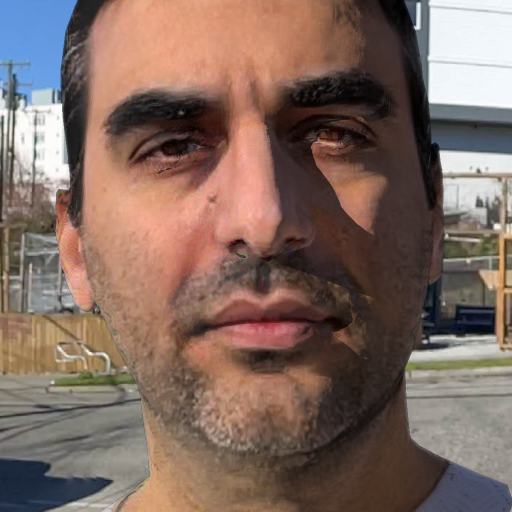} %
\caption{Less shinier}
\end{subfigure}
\begin{subfigure}[t]{0.24\linewidth}
\includegraphics[width=\linewidth]{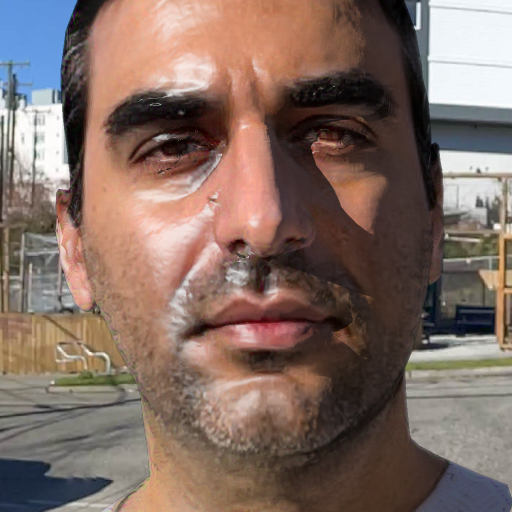} %
\caption{More shinier}
\end{subfigure}

\caption{\textbf{Adjusting the specularity}. We can change the the specular properties of the face, making the face less shinier (a, c) or more shinier (b, d).}
\label{fig:qualitative-shiny}

\vspace{-18pt}

\end{center}
\end{figure}

% \subsection{Skin Reflectance Editing}
\paragraph{Skin Reflectance Editing}
We are also able to edit the reflectance components of the subject. As shown in Figure~\ref{fig:teaser}e, we can adjust the optimized albedo to add freckles, stickers, or other textures that realistically interact with reflections, shadows, and other elements of scene lighting, or we can adjust the specular properties of the face, making the face more or less shinier, as shown in Figure~\ref{fig:qualitative-shiny}.
%Please see more results in the supplemental materials.

\section{Conclusion and Discussions}
\label{sec:conclusion}

In this paper, we propose SunStage, a lightweight and practical facial capture, rendering, and editing system that can serve as a minimalist light stage. With a video of an individual rotating in-place under the sun, our system reconstructs a physical model of the subject and the scene lighting, which enables us to relight the subject with realistic reflections and cast shadows. Our system allows arbitrary lighting and reflectance control in the reconstructed physical space, which can be rendered to produce photo-realistic results. We demonstrate several applications such as editing skin reflectance, relighting, and view synthesis.

\paragraph{Limitations} Our system inherits the limitations of morphable face models and is unable to model hair, teeth, or clothing geometry, beyond slight deformations. Additionally, certain regions which are seen under constant shadow or specular reflection (and therefore have no cues on reflectance or albedo) are sometimes unable to be decomposed accurately into separate reflectance and lighting components. Visualization and further discussions of the system's limitations are provided in the supplemental material.

%%%%%%%%% REFERENCES
{\small
\bibliographystyle{ieee_fullname}
\bibliography{main}
}

\clearpage
\appendix

\section{Formulation Details}
We expand the diffuse and specular contribution derivations below:

\paragraph{Diffuse contribution}
The diffuse contribution $L^d_o$ is given by the diffuse terms of Equation~5 in the main paper:
%\ref{eqn:render}:
\begin{align}
    &L_o^d(x) \nonumber\\
    &=\sum_{\omega_i} V(x, \omega_i) L_i^\text{amb}(\omega_i) \odot R^d(x, \omega_i) \left( \omega_i \cdot n(x) \right) \Delta\omega_i \nonumber\\
    &+ \sum_{\omega_i} V(x, \omega_i) L_i^\text{sun}(\omega_i) \odot R^d(x, \omega_i) \left( \omega_i \cdot n(x) \right) \Delta\omega_i\\
    &= \sum_{\omega_i} V(x, \omega_i)L_i^\text{amb}(\omega_i) \odot \frac{a(x)}{\pi}(\omega_i \cdot n(x)) \Delta\omega_i \nonumber\\
    &+ V(x, p^\text{sun}) k^\text{sun}[1, 1, 1] \odot \frac{a(x)}{\pi} (p^\text{sun} \cdot n(x)) \Delta p^\text{sun}\\
    &\approx \sum_{\omega_i} L_i^\text{amb}(\omega_i) \odot \frac{a(x)}{\pi}(\omega_i \cdot n(x)) \Delta\omega_i \nonumber\\
    &+ V(x, p^\text{sun}) k^\text{sun}[1, 1, 1] \odot \frac{a(x)}{\pi} (p^\text{sun} \cdot n(x)) \Delta p^\text{sun}\,
\end{align}
where $a(x)$ is the albedo at point $x$, $k^\text{sun}$ is the (optimized) sun intensity, and $p^\text{sun}$ is the (optimized) sun direction. The sun is modeled as a directional light source, so the second summation can be simplified to a single term (i.e. only in the direction of $p^\text{sun}$). $E \in \mathbb{R}^{16 \times 32 \times 3}$ is the high-dynamic-range (HDR) environment map. We ignore the visibility term for the ambient lighting during optimization, as it is computationally intensive to compute the visibility for all light directions and the ambient intensity is much weaker than that of the sun.

\paragraph{Specular contribution}
The specular contribution $L^s_o$ at each pixel is given by the specular term of the sun, see Equation~5 in the main paper 
% ~\ref{eqn:render} 
(recall that we ignore the specular term of the ambient due to its weak contribution):
\small
\begin{align}
    &L^s_o(x, \omega_o) \nonumber\\
    &= \sum_{\omega_i} V(x, \omega_i) L_i^\text{sun}(\omega_i) \odot R^s(x, \omega_i, \omega_o) \left( \omega_i \cdot n(x) \right) \Delta\omega_i \\
    &= V(x, p^\text{sun}) k^\text{sun} [1, 1, 1] k_s \frac{s + 2}{2 \pi} \left(h(p^\text{sun}, \omega_o) \cdot n(x) \right)^s \nonumber\\
    &\left( p^\text{sun} \cdot n(x) \right) \Delta p^\text{sun} \,
\end{align}
\normalsize
where we have substituted Equation~2 in the main paper 
% \ref{eqn:brdf} 
and reduced the summation to just one term at $p^\text{sun}$ (i.e. $L_i^\text{sun}$ is $0$ elsewhere).

\section{Optimization Details}

\subsection{Coarse optimization}
The pose loss, $L_\text{pose} = L_\text{mask} +  L_\text{lmk}$, is optimized using an ADAM optimizer \cite{kingma2014adam} with the following learning rates for different parameters:
{
\begin{center}
\begin{tabular}{l|l}
$(\beta, \theta_j, \psi_j)$ & $1e{-4}$\\
$S$ (Object scale) & $1e{-2}$\\
$T_j$ (Object translation) & $1e{-2}$
\end{tabular}\\ 
\end{center}
}
\noindent We optimize for a total of 2,000 epochs on a single NVIDIA RTX 2080 Ti GPU.
Each epoch consists of one optimization step for all training images. The coarse optimization takes $2.5$ hours to converge for a sequence that contains $200$ images with resolution $224 \times 224$.

To account for the fact that our geometry does not explicitly model hair and clothing, our mask loss $L_\text{mask}$ consists of a foreground mask loss $L_\text{mask\_foreground}$ which corresponds to skin regions and a background mask loss $L_\text{mask\_background}$ which corresponds to background pixels.

\begin{equation}
    L_\text{mask} = L_\text{mask\_foreground} + L_\text{mask\_background}
\end{equation}
where the foreground mask loss is enforcing $1$s in the skin region and the background mask loss is enforcing $0$s in the background region. Both are L2 losses.

% is only defined in regions which correspond to skin or background pixels. Pixels corresponding to clothing or hair are ignored, as to avoid optimizing the geometry to account for them. 
We compute the background mask using the matting model from RVM~\cite{lin2022robust}, and the skin mask from a modified version\footnote{https://github.com/zllrunning/face-parsing.PyTorch} of BiSeNet \cite{yu2018bisenet,yu2021bisenet}. The landmarks used for the landmark loss are obtained from HRNet~\cite{wang2020deep}.

\subsection{Photometric optimization}
The final loss:
\begin{align}
    L = &\lambda_\text{mask}L_\text{mask} + \lambda_\text{lmk} L_\text{lmk} + \lambda_E L_{E} + \nonumber \\
    &\lambda_{E\text{s}} L_{E\text{s}} + \lambda_{\text{VGG}} L_{\text{VGG}} + \lambda_{\text{photo}} L_{\text{photo}}
\end{align}
is optimized using ADAM for 4,000 epochs. The initial learning rates are\\
{
\begin{center}
\begin{tabular}{l|l}
$(\beta, \theta_j, \psi_j)$ & $1e{-4}$\\
$\Delta X$ & $1e{-4}$\\
$S$ & $1e2$\\
$T_j$ & $1e2$\\
$a$ & $1e{-2}$\\
$s$ & $1e{-2}$\\
$k_s$ & $1e{-2}$\\
$p^\text{sun}$ & $1e{-3}$\\
$E$ & $1e{-3}$\\
$k^\text{sun}$ & $1e{-3}$
\end{tabular}\\ 
\end{center}
}

\noindent Every 1,000 epochs, we decay the learning rate of all parameters to $10\%$ of their previous values. The optimization runs on a single NVIDIA RTX 2080 Ti GPU and takes $2$ days to converge for a sequence of $200$ images of resolution $224 \times 224$. The VGG loss $L_{VGG}$ in the photometric optimization follows the implementation of pix2pixHD~\cite{wang2018pix2pixHD}.

\section{Capture Details}

Our video sequences are captured by an iPhone 12 Pro Max or an iPhone 13 Pro Max. While we have validated that our method works equally well with keypoint-based pose estimation techniques, we opt for using the phone's integrated visual-inertial SLAM system, since we found empirically that it produces more reliable camera orientations. In practice, to capture our sequences, we use CamTrackAR\footnote{https://fxhome.com/product/camtrackar}, an app that captures synchronized camera intrinsics\footnote{since the iPhone camera uses optical image stabilization, and therefore the principal point and focal lengths vary as the camera moves}, poses, and video frames. 

During capture, we impose constraints on the subject expression to account for the limitations of the face morphable model we use for optimization, e.g. not modeling the teeth (see more discussions in Section~\ref{supp-limitation}).
Prior to capture, our subjects are instructed to try to maintain a constant expression, to face forward, and to rotate in place. Naturally, it is nearly impossible to remain perfectly centered and to keep a constant pose and expression, but fortunately, our formulation is tolerant to variations in both. In fact, our reconstructed geometry can typically model subtle expression variations like smiles and twitches. Still, to reduce noise in optimization, we filter out the input frames in which the subject has an open mouth or is blinking, or frames which contain significant glare (i.e., when the camera is facing the sun).

While a single rotation provides sufficiently many constraints on the shape of the face (through shadows and specular reflections), the quality of the geometry and texture at the boundaries of the face (i.e., at the edge of the jaw and the side of the face) can be improved by capturing additional frames where the camera is rotating independently from the face. In these cases, we capture a full 360 degree rotation, then stop, and continue rotating the camera along a 15 degree arc back and forth. When using this capture technique, we typically sample around $200$ frames from the full video for optimization -- $100$ from the initial 360 degree rotation, and $100$ from the arc sequence. Since all images from the arc sequence provide similar photometric constraints (i.e. they do not give strong constraints on material properties, lighting, or normal, given that the lighting conditions are identical), and in order to avoid degenerate optimization, we only optimize for at most five arc images per epoch.

\section{Evaluation Details}
In this section, we describe how we compare SunStage with ablated variants and baselines, for both the tasks of relighting and view synthesis. Since different baseline approaches make different assumptions, we clarify the necessary adaptations made for fair comparisons. We also provide analysis of the results. 

\subsection{Held-out frames}
In addition to capturing a single 360-degree rotation, we capture two testing sequences: one for evaluating view synthesis, and one for evaluating relighting quality.

Immediately after completing the capture, the subject hands the camera to another person, who captures a multi-view video of the subject (i.e., translating and rotating the camera to capture the face from different viewpoints). The subject remains still during this process. Of these captured frames, the first (which is typically the frame facing the subject head-on), is included in the ``training" sequence (and is used as the input for other methods that only operate on single frames), and the remainder are held out as testing images for view synthesis. For the purposes of evaluation, we assume the subject is entirely stationary for this portion of the capture.

Additionally, we ask the subject to capture a second sequence, either in a different location, at a different time of day, or at a different relative angle from the sun. We use these examples to evaluate our method's ability on relighting. Both sequences are reconstructed separately, without using a shared model for the subject's geometry, reflectance, or appearance. For evaluation, we swap the estimated lighting conditions between the two models, i.e., we render an image using the estimated sun direction and environment map (as well as a given frame's pose and expression) from sequence A and the subject geometry and materials from sequence B, and compare the result to the corresponding real image from sequence A. These target poses, expressions, and lighting conditions are shared by all ablations and baseline experiments.  

\subsection{Metrics}

To quantitatively compare these methods, we compute relighting and view synthesis errors measured in Peak Signal-to-Noise Ratio (PSNR), Similarity Index Measure (SSIM) \cite{wang2004image}, and Learned Perceptual Image Patch Similarity (LPIPS) \cite{zhang2018perceptual}.
As Table~1 in the main paper
% \ref{tbl:quant} 
shows, SunStage achieves the best performance in both relighting and view synthesis across all three error metrics.

For the task of relighting, we composite all methods' results onto the ground-truth frames using skin masks extracted using a modified version\footnote{https://github.com/zllrunning/face-parsing.PyTorch} of BiSeNet \cite{yu2018bisenet,yu2021bisenet}, since our method does not relight non-skin regions, e.g. hair and clothes. The composited result is used for comparison against the ground-truth images.

\subsection{Baseline Comparisons}

In the following paragraphs, we provide the details about how we train and test the baseline models, as well as our analysis of these baseline results.

\paragraph{DECA}
As a very na\"ive baseline, we use DECA~\cite{feng2021learning}, a single-image face reconstruction method, for the tasks of novel-view synthesis and relighting. From a single image, DECA predicts facial geometry, spherical harmonic lighting, and albedo. For view-synthesis, we run DECA on two images (separately) to get two sets of albedo, geometry, and lighting. To render an image from a new viewpoint, we simply swap the shape code, expression code and albedo from one image to the other. While this gives DECA a significant advantage, since the lighting and pose are estimated directly from the ground-truth frame, we find in practice that the rendered results seldom resemble the ground-truth images. Additionally, although DECA contains a deformation map to model fine details, we find in practice that this seldom accurately models subject-specific geometry details such as wrinkles. One reason for the poor performance at this task is the orthographic assumption. As shown in Figure~6 in the main paper,
% \ref{fig:ablation-2-stage}, 
DECA's pose and shape estimates significantly deteriorate upon introduction of strong perspective effects. 

For the task of relighting, we similarly run DECA on a pair of images, and swap the lighting conditions between the two. We find that since DECA assumes a Lambertian model, the resulting images are far from photorealistic. 

\paragraph{GCFR}
GCFR~\cite{hou2022face} is a single-image relighting method that aims to handle hard shadows in new lighting scenarios. It predicts a shadow mask from an estimated depth map of the face. We use the pretrained model from GCFR for the baseline comparison.

Given the input and target image, we first use GCFR's Shadow Mask Estimation module to estimate the shading map, then we estimate the albedo map from the input image using GCFR's hourglass network (albedo decoder). To render the target relit image, we compose the \textit{target shading map} (advantageous to the baseline) with input albedo map following Equation~6 in~\cite{hou2022face}.

However, GCFR fails to accurately relight images even with the target shading map. Its hourglass network is not able to estimate a good albedo from the input image --- the estimated albedo often has shadows and specular highlights baked in. This is likely because the training dataset of GCFR does not contain enough images with hard shadows.

\paragraph{DPR}
DPR~\cite{zhou2019deep} is a single-image learning system that operates entirely in the 2D image space. We use the pretrained model from DPR as a baseline for the task of relighting.

For simplicity, we evaluate DPR only on same-environment relighting. In other words, we ask DPR to render the same environment as the input sequence, but under different incident sun angles. To render a relit image, we first run DPR to estimate the spherical harmonics coefficients of the input environment map, then rotate these towards the lighting of the target image, and finally feed those SH coefficients to perform relighting. DPR additionally requires the input and output to be spatially aligned. To facilitate this, we use the geometry estimated by SunStage as a proxy for reprojecting an input frame into the pose of the target frame. 

We find that DPR is unable to accurately relight images, failing to synthesize accurate shadows and specularity, largely as a result of the lack of explicit 3D reasoning of the subject.

\paragraph{Total Relighting}
Total Relighting~\cite{pandey2021total} (TR) is a state-of-the-art single-image method trained on high-quality light state data, achieving impressive shading and specular highlights.

We use Total Relighting as a baseline for relighting. As input, we provide the same reprojected images as we do for DPR, and provide our target environment maps (the same ones used for evaluating our method) as the target lighting conditions. 

Since TR does not model cast shadows explicitly, it often has difficulty removing cast shadows from input images, and does not produce accurate cast shadows in the relit images.
Additionally, Total Relighting produces color tones that do not match the target images. We suspect this occurs as a result of a number of factors: (1) there is an inherent ambiguity in (particularly single-frame) decomposition of lighting and albedo, and Total Relighting may simply be decomposing the two differently when compared to our method, and (2) the lighting maps provided for our quantitative relighting tasks (i.e. those in the main paper, not the HDRI renderings shown in the supplement) are the result of our method's decomposition, and therefore may not match exactly the characteristics of the HDRI images used for training TR. 
We use the pretrained Total Relighting model (from the authors) to run inference on our images.

\paragraph{NLT}
Like SunStage, NLT~\cite{zhang2021neural} is a ``test-time optimization'' approach that learns a subject-specific appearance model from multiple observations of the same subject.

We use NLT as a baseline for both relighting and view synthesis. Since NLT expects that the incoming light directions and viewing directions are known and a geometry proxy is provided, we train NLT on the input images, along with the camera poses, sun directions, and face geometry estimated by SunStage. At test time, we query the trained NLT with novel sun directions for relighting and novel viewpoints for view synthesis.

We find that for both tasks, NLT produces less sharp specular highlights and an overall less accurate rendering than SunStage. NLT produces blurry results and ghosting shadows, likely due to the discrepancy in the number of images used for training and the number of images typically captured by a light stage (NLT has been shown to work on $300\times50=15,000$ images, as opposed to ours that uses only $200$).

\paragraph{NextFace}
Similar to SunStage, NextFace~\cite{dib2021towards} is an optimization-based face reconstruction method. It learns to decompose the input image into shape, lighting, and material properties from multiple observations of the same subject. 

We evaluate NextFace on both relighting and view synthesis. We first train NextFace on our training images to estimate the shape, lighting, and material properties. For the test set, we train another NextFace model to get the estimated lighting as the target for relighting, and shape parameters as the target for view synthesis. To render the final image for relighting, we use the lighting at test time, and shape and materials at training time. For view synthesis, we use the shape at test time, and lighting and material at training time. We use the ray tracer in NextFace as the renderer.

We find that for both tasks, NextFace fails to model self-cast shadows. This is due to the lighting formulation it adopts. NextFace uses spherical harmonics (SH) to model the scene lighting, which is unlikely to model high-frequency lighting such as the hard sunlight.

\subsection{Ablations}

In this section, we examine different ablated versions of our method.

\begin{figure*}[t!]
\begin{center}
\includegraphics[width=1.00\linewidth]{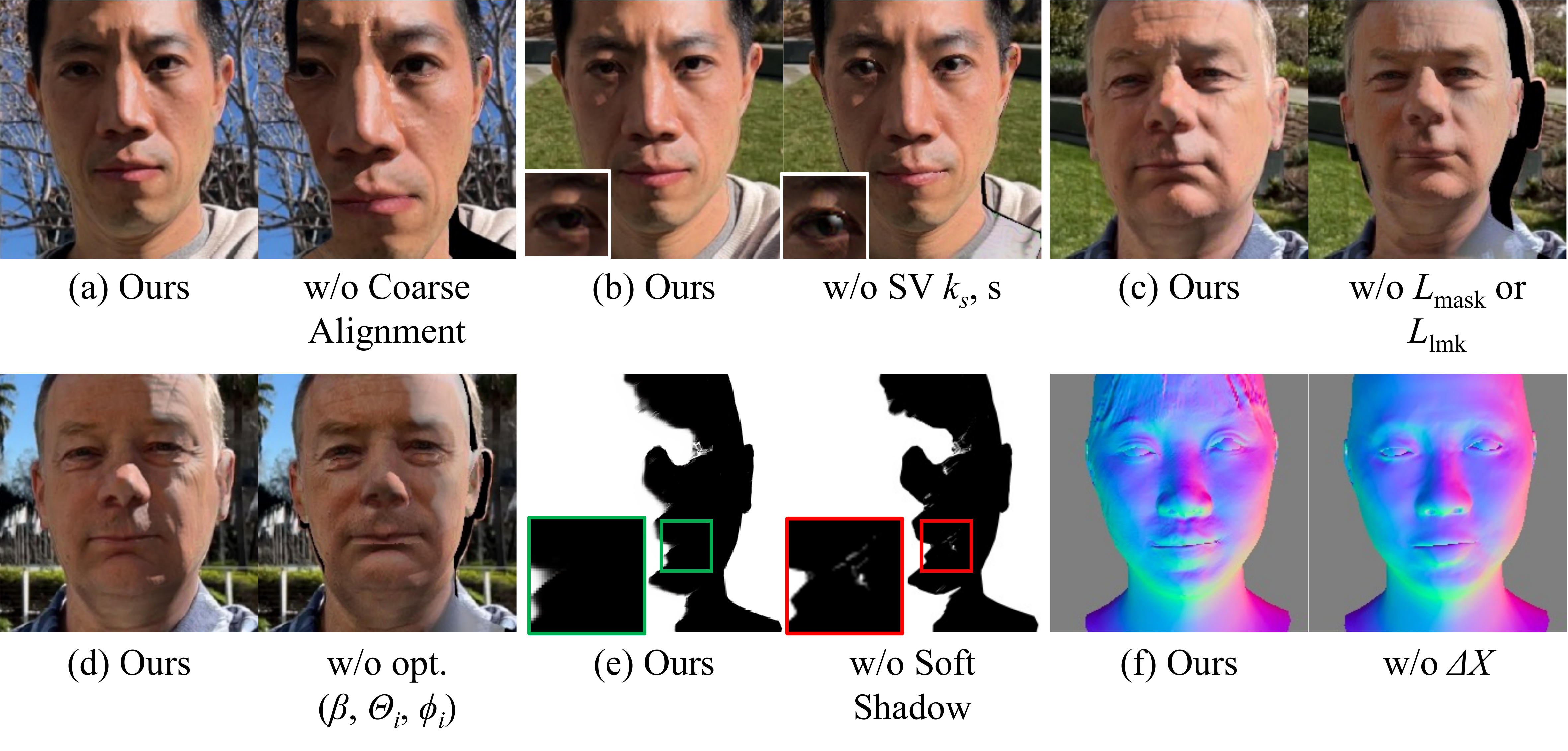}
\end{center}
\vspace{-6mm}
\caption{%
Visualization of common artifacts produced by the ablated versions of SunStage.
}
\vspace{-2mm}
\label{fig:supp_ablations}
\end{figure*}

\paragraph{Ours w/o coarse}
In this ablation, we directly optimize for all parameters without the coarse alignment stage.
In practice, we find that optimization seldom converges to a reasonable solution due to the ill-posed nature of our optimization problem:
different combinations of geometry, reflectance, lighting, and camera poses may lead to the same observed image.

As Figure~\ref{fig:supp_ablations}a shows, the optimization result, without coarse alignment, often gets trapped in local optima. Quantitatively, this ablated version of SunStage falls far behind the full model.

\paragraph{Ours without spatially varying (SV) specular coefficients $k_s$, $s$}
This ablation study explores how spatially-varying shininess (i.e., the $k_s$ and $s$ parameters of our reflectance model) is critical for recovering a photorealistic facial reflectance model.
Since different face regions possess different shininess factors, using a global $k_s$ and $s$ value leads to an averaged solution, where no area is estimated to be strongly shiny to avoid large re-rendering errors.

Another artifact we observe is that the eyeballs are estimated to have overly wide specular lobes (similar to that of the skin), as shown in Figure~\ref{fig:supp_ablations}b.
Quantitatively, this variant performs worse than our full model but still achieves reasonable errors (ranked the third best for relighting). This is likely due to the fact that the specular component is numerically insignificant. 

\paragraph{Ours w/o $L_\text{mask}$, $L_\text{lmk}$} 
This model variant ablates the contribution of mask loss and landmark loss in the second (i.e., photometric) stage of our optimization. For this variant, we preserve the first stage solution and disable the mask and landmark losses for the second stage of optimization.

Similar to the ``w/o coarse'' ablation, this model variant solves a less constrained optimization problem than our full model does, e.g., without the facial keypoints, there is no constrains on the mouth image pixels to align with the geometry corresponding to the mouth. 
We observe similar qualitative (Figure~\ref{fig:supp_ablations}c) and quantitative results (Table~1 in the main paper
% \ref{tbl:quant}
) as in the ``w/o coarse'' ablation.

\paragraph{Ours w/o $L_\text{mask}$}
Similar to the previous ablation, we turn off $L_\text{mask}$ only in the second stage of our optimization.
This ablation suffers from alignment issues as seen before, and therefore we skip its visualization in Figure~\ref{fig:supp_ablations}.
Quantitatively, as shown in Table~1 in the main paper
% \ref{tbl:quant}
, this ablation outperforms ``w/o $L_\text{mask}$, $L_\text{lmk}$'' by having more optimization constraints from $L_\text{lmk}$ but still underperforms our full model by a large margin.

\paragraph{Ours w/o opt.\ $(\beta, \theta_i, \phi_i)$}
In this experiment, we preserve the initialized shape provided by DECA without refining it.
Since the images captured under our setup are selfies, the perspective effects are not accounted for by DECA, which assumes an orthographic camera model. As such, the shape estimated by DECA is not well-aligned with our input images. As shown in Figure~\ref{fig:supp_ablations}d and Table~1 in the main paper
% \ref{tbl:quant}
, our shape optimization strategy improves the initialized DECA shape.

\paragraph{Ours w/o soft shadow}
In this variant, instead of doing the soft comparison as Equation~8 in the main paper
% \ref{eq:soft} 
states, we use a hard z-buffer comparison in producing the shadow maps.
Although Table~1 in the main paper
% \ref{tbl:quant} 
shows that this ablated version of our model achieves reasonable quantitative performance, as Figure~\ref{fig:supp_ablations}e demonstrates, using a hard comparison produces spurious shadows, especially when the sun is at grazing angles. Additionally, the optimized shadows (and the sun position) are less accurate, which is likely due to the instability in optimization as the gradients are not continuous for the hard shadow comparison formulation.
 
\paragraph{Ours w/o $\Delta X$}
We also explore the quality of our method without optimizing for a displacement map. As Figure~\ref{fig:supp_ablations}f illustrates, this ablation is unable to model geometric details such as wrinkles and pores. Consequently, such effects are baked into the albedo, causing artifacts in applications such as relighting and material editing. Additionally, this ablation produces blurrier renderings, since the high-frequency appearance change is harder to be explained by other factors such as reflectance and lighting.

\begin{figure}
\begin{center}
\centering
\begin{subfigure}[b]{0.32\linewidth}
 \centering
 \includegraphics[width=\linewidth]{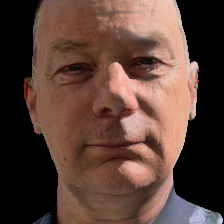}\\
 \includegraphics[width=\linewidth]{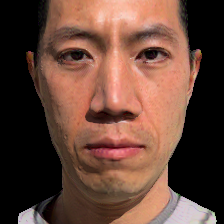}\\
 \caption{Blinn-Phong}
\end{subfigure} \hfill %
\begin{subfigure}[b]{0.32\linewidth}
 \centering
 \includegraphics[width=\linewidth]{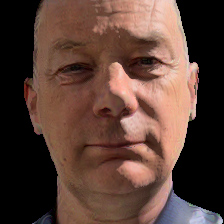}\\
 \includegraphics[width=\linewidth]{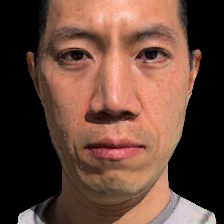}\\
 \caption{Microfacet~\cite{walter2007microfacet}}
\end{subfigure} \hfill %
\begin{subfigure}[b]{0.32\linewidth}
 \centering
 \includegraphics[width=\linewidth]{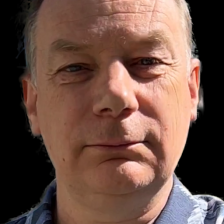}\\
 \includegraphics[width=\linewidth]{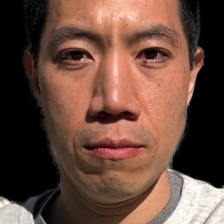}\\
 \caption{Ground truth}
\end{subfigure}
\end{center}
\caption{
Comparison on different reflectance models. A more complex model~\cite{walter2007microfacet} does not significantly improves visual quality. On the other hand, it is hard to optimize and introduces instability in training.
}
\label{fig:supp_sp}
\end{figure}
\begin{figure*}[t!]
\begin{center}

\includegraphics[width=\linewidth]{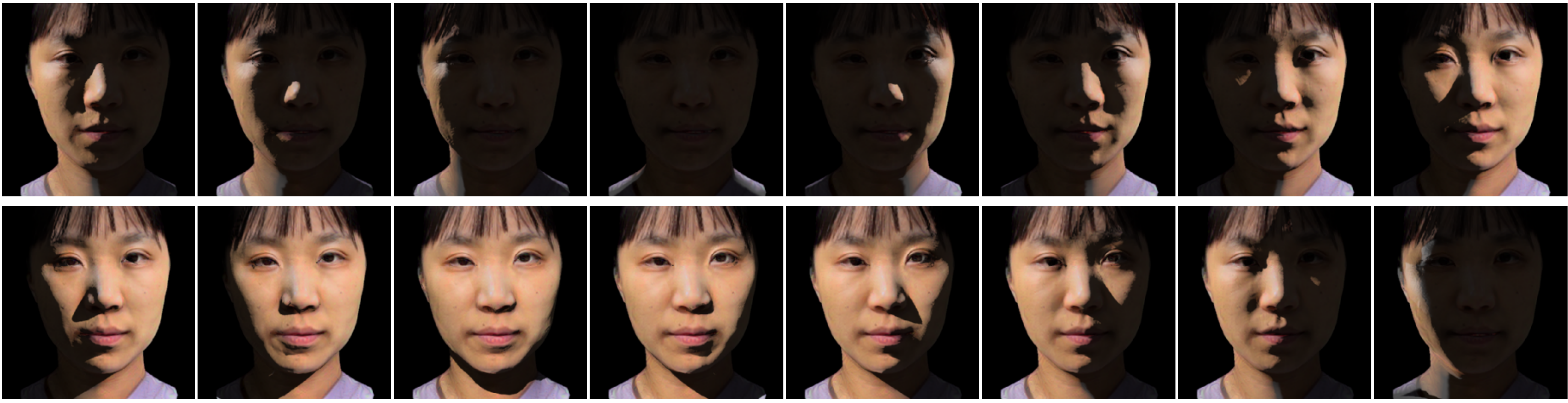}

\caption{\textbf{Synthetic OLAT}. By rendering the recovered face with a single distant light source (where geometry artifacts are exposed), we can simulate the One-Light-at-A-Time (OLAT) data that was only possibly captured with a light stage.}
\label{fig:app-olats}

%\vspace{-7mm}

\end{center}
\end{figure*}

\paragraph{Ours w/ microfacet reflectance model}
Finally, we change the Blinn-Phong reflectance model to a more complex microfacet reflectance model~\cite{walter2007microfacet}. As shown in Figure~\ref{fig:supp_sp}, microfacet model produces comparable specular highlights with that from the Blinn-Phone model. The microfacet model~\cite{walter2007microfacet} describes the complicated light paths that depend on incoming light direction, surface normal and material properties. The gradients on these parameters, which contribute to multiple terms in the equation, are more noisy comparing to the simple Blinn-Phong reflectance model. Using the same optimization scheme as in Blinn-Phong, we find it impossible for the scene parameters converge to a reasonably steady state. Therefore, we turn off the specular highlights for the first $100$ epochs to reduce the parameter entanglement, and start optimizing all variables in the microfacet model once the light (i.e., the sun) position is converged. We observe little visual quality difference between using microfacet and Blinn-Phong reflectance models, while the former involves a much more unstable and difficult optimization scheme. SunStage thus uses the simple Blinn-Phong reflectance model.
% Furthermore, the complexity of microfacet model makes it hard to optimize and introduces instability in training. It often requires additional optimization schemes to converge.

\section{Additional Results}

In Figure~\ref{fig:supp_tr_circus_1}, Figure~\ref{fig:supp_tr_circus_2}, Figure~\ref{fig:supp_tr_qm_1} and Figure~\ref{fig:supp_tr_qm_2}, we show more comparisons with Neural Video Portrait Relighting (NVPR)~\cite{zhang2021neuralvideo} and Total Relighting (TR)~\cite{pandey2021total}. Both NVPR and TR are image based relighting methods which leverage the priors learned from light stage data. 
At test time, the model takes in an arbitrary input image and a target HDR environment map, and generates a relit result. We find that neither of the baselines fully preserves the identity of the subject, changing facial geometry or missing some of the detailed reflectance properties (e.g. accurate specular highlights) that are unique to each individual subject. Both NVPR and TR also leave harsh traces on the relit results at the locations where the shadow boundary exists in the input image (see Figure~\ref{fig:supp_tr_circus_1} row 1 and row 3). This is likely a result of the lack of such images (i.e., with harsh shadows) in the training dataset.

\begin{figure}
\begin{center}
\includegraphics[width=0.4\linewidth]{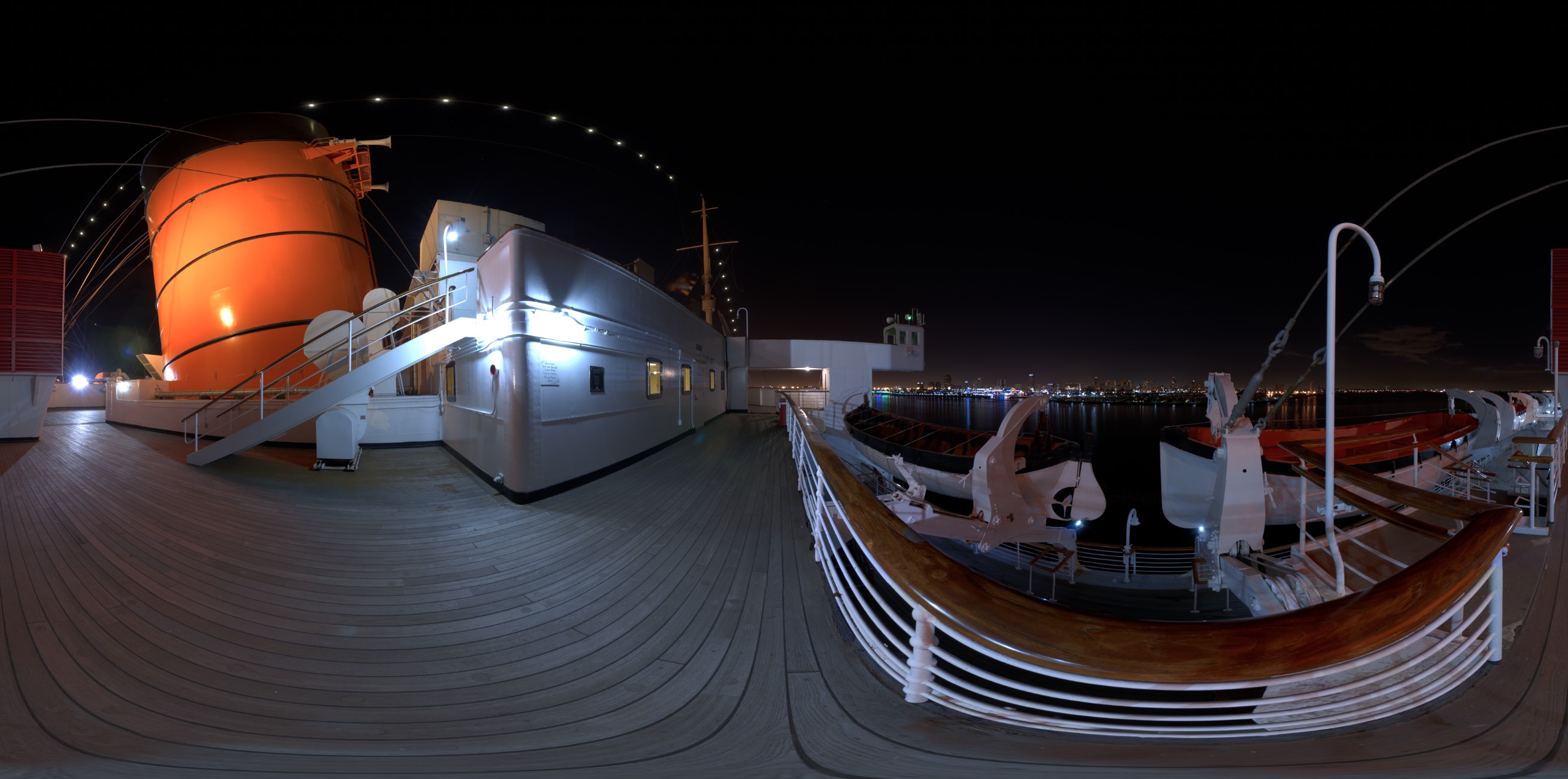}
\includegraphics[width=0.4\linewidth]{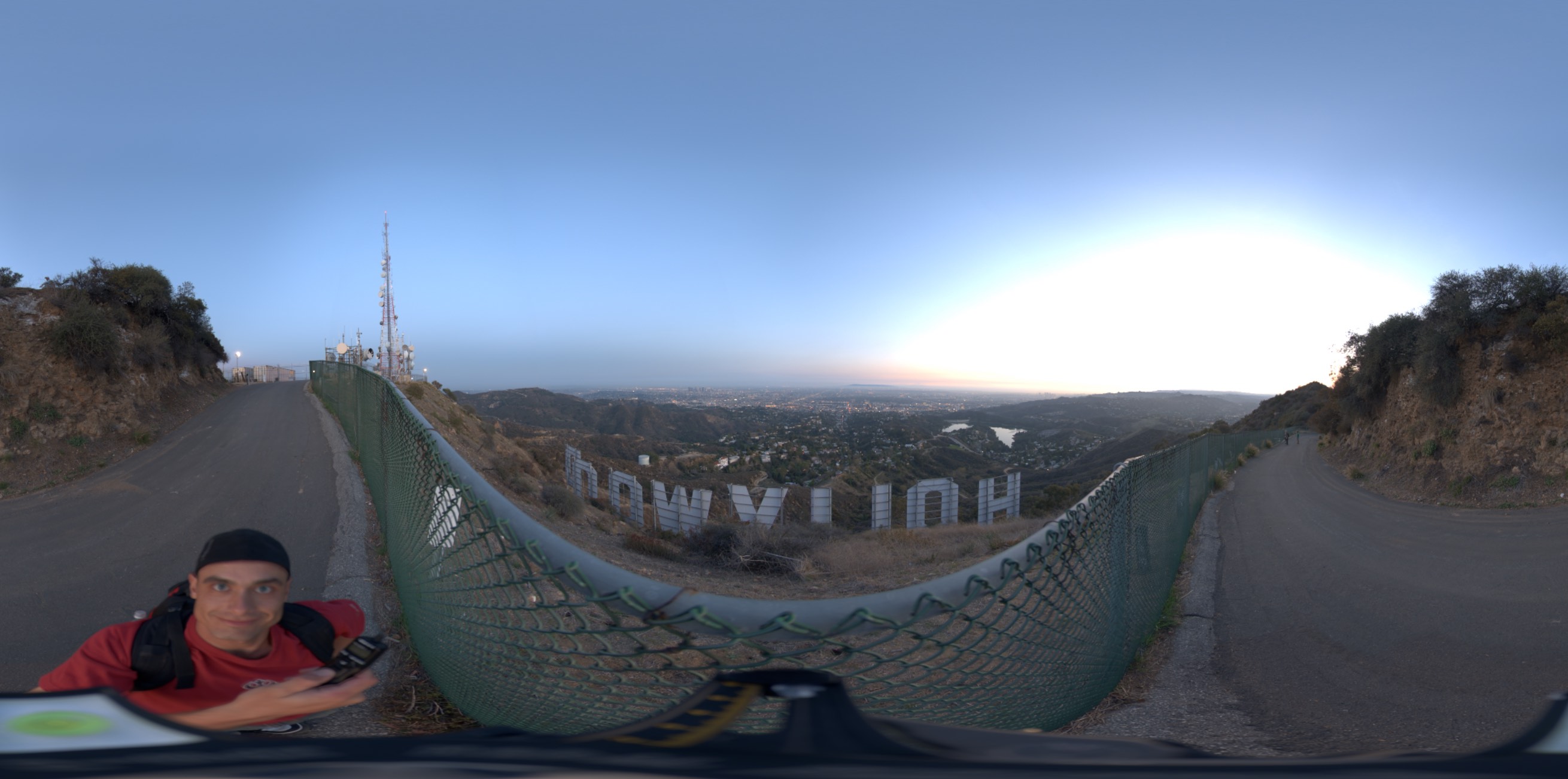}
\end{center}
\caption{
Target lighting: Queen Mary (left) and Hollywood (right) for the following comparisons with Neural Video Portrait Relighting and Total Relighting. Both environment maps are tone mapped for visualization.
}
\label{fig:supp_tr_envmap}
\end{figure}

\begin{figure}
\begin{center}

\centering
\begin{subfigure}[b]{0.24\linewidth}
 \centering
 \includegraphics[width=\linewidth]{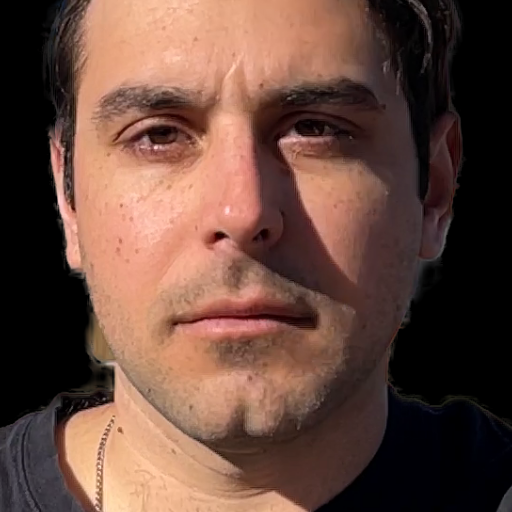}\\
 \includegraphics[width=\linewidth]{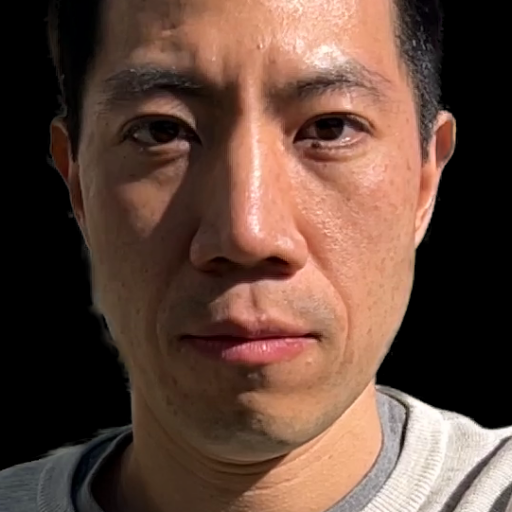}\\
 \includegraphics[width=\linewidth]{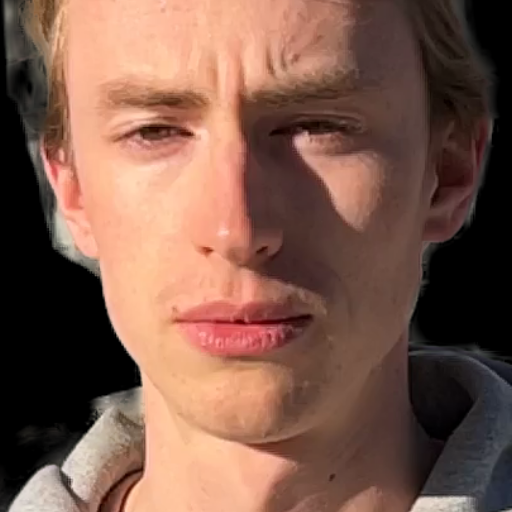}\\
 \includegraphics[width=\linewidth]{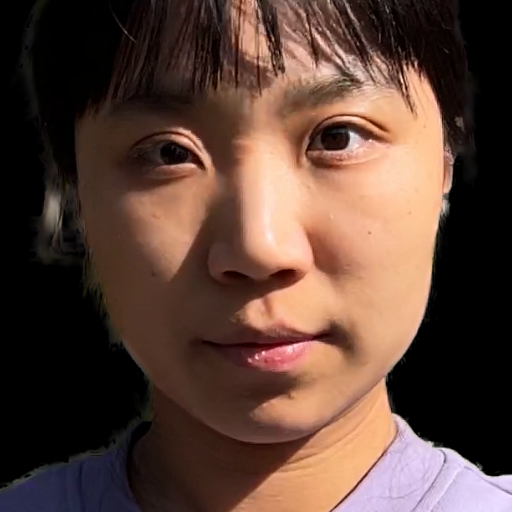}\\
 \caption{Input}
\end{subfigure} \hfill %
\begin{subfigure}[b]{0.24\linewidth}
 \centering
 \includegraphics[width=\linewidth]{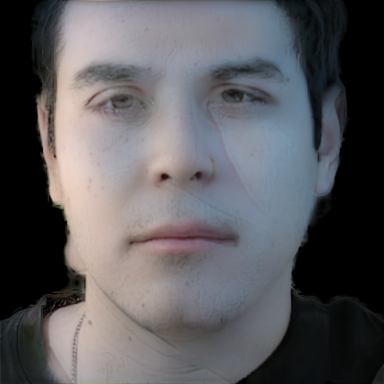}\\
 \includegraphics[width=\linewidth]{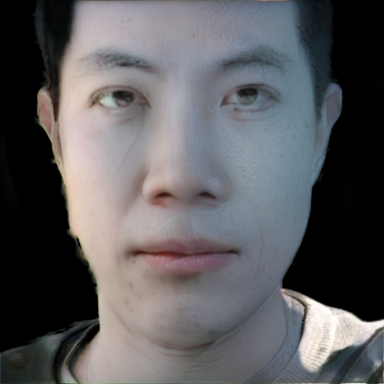}\\
 \includegraphics[width=\linewidth]{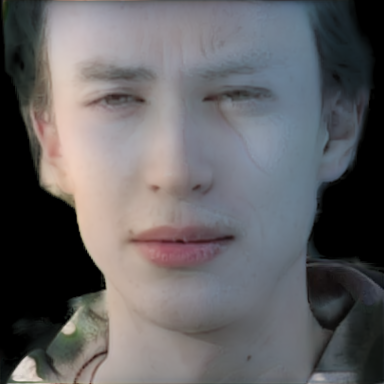}\\
 \includegraphics[width=\linewidth]{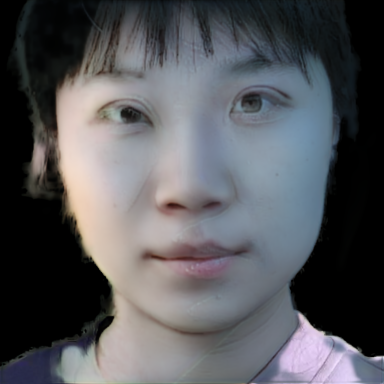}\\
 \caption{NVPR~\cite{zhang2021neuralvideo}}
\end{subfigure} \hfill %
\begin{subfigure}[b]{0.24\linewidth}
 \centering
 \includegraphics[width=\linewidth]{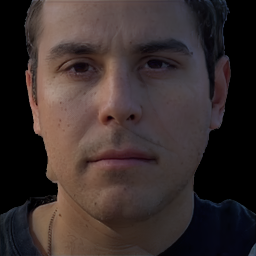}\\
 \includegraphics[width=\linewidth]{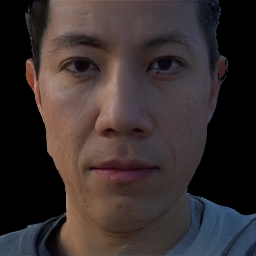}\\
 \includegraphics[width=\linewidth]{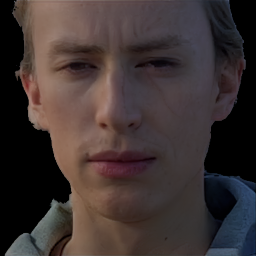}\\
 \includegraphics[width=\linewidth]{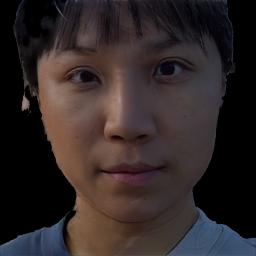}\\
 \caption{TR~\cite{pandey2021total}}
\end{subfigure} \hfill %
\begin{subfigure}[b]{0.24\linewidth}
 \centering
 \includegraphics[width=\linewidth]{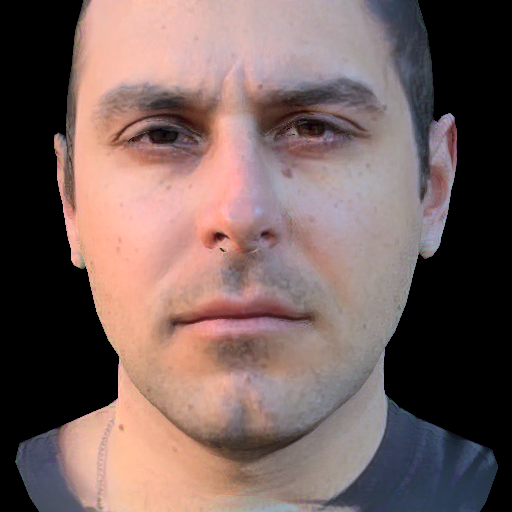}\\
 \includegraphics[width=\linewidth]{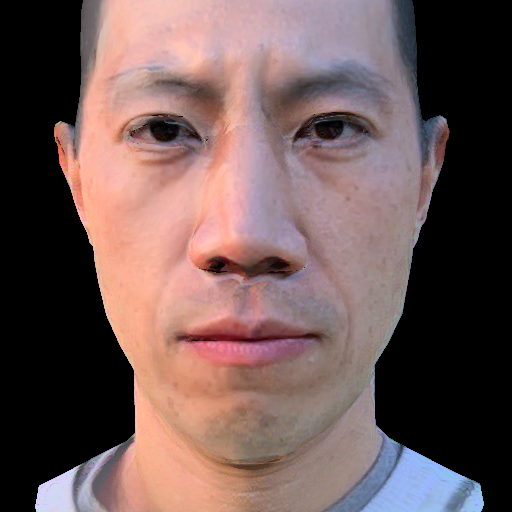}\\
 \includegraphics[width=\linewidth]{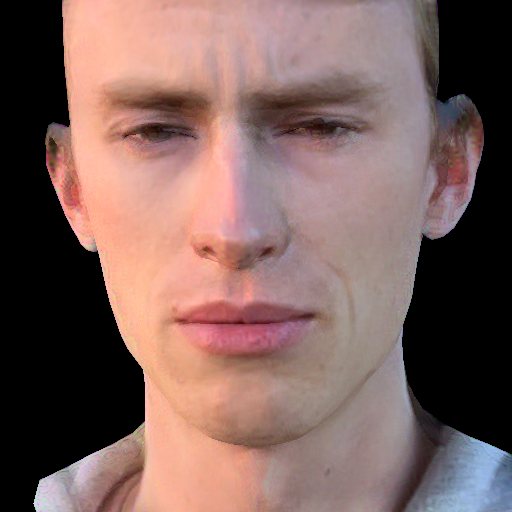}\\
 \includegraphics[width=\linewidth]{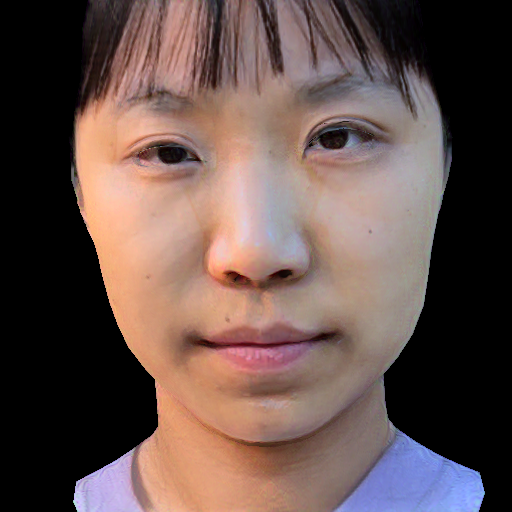}\\
 \caption{Ours}
\end{subfigure}
\end{center}
\caption{
Comparison with Neural Video Portrait Relighting and Total Relighting on the target lighting ``Hollywood''. Both NVPR and TR leverage face priors by training on large-scale light stage data. While being able to generalize to an arbitrary input, these methods can not model the individual skin reflectance properties. Both NVPR and TR also leave traces of visible shadow boundaries from the input images.
}
\label{fig:supp_tr_circus_1}
\end{figure}

\begin{figure}
\begin{center}
\centering
\begin{subfigure}[b]{0.24\linewidth}
 \centering
 \includegraphics[width=\linewidth]{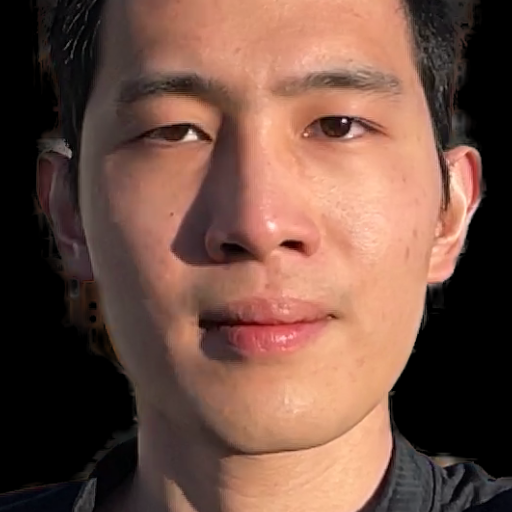}\\
 \includegraphics[width=\linewidth]{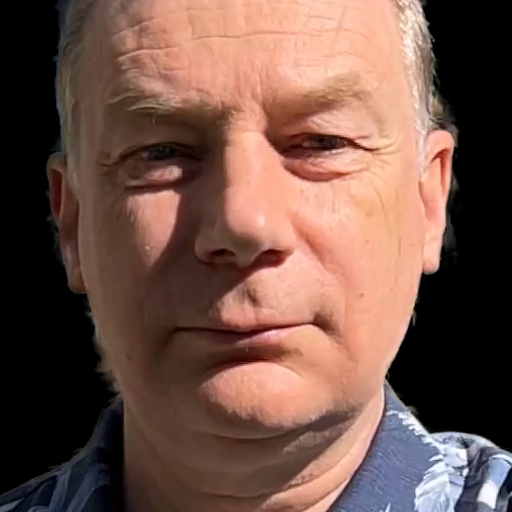}\\
 \includegraphics[width=\linewidth]{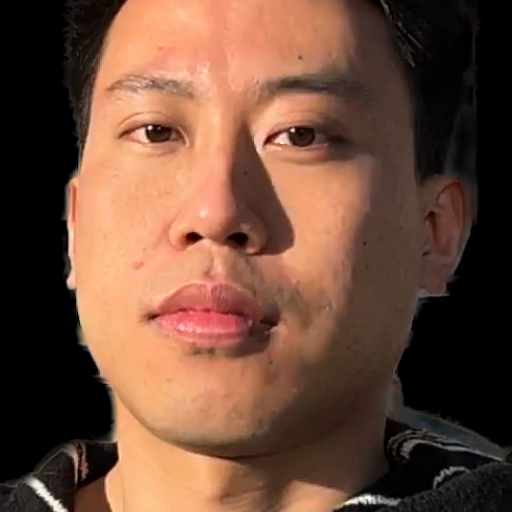}\\
 \includegraphics[width=\linewidth]{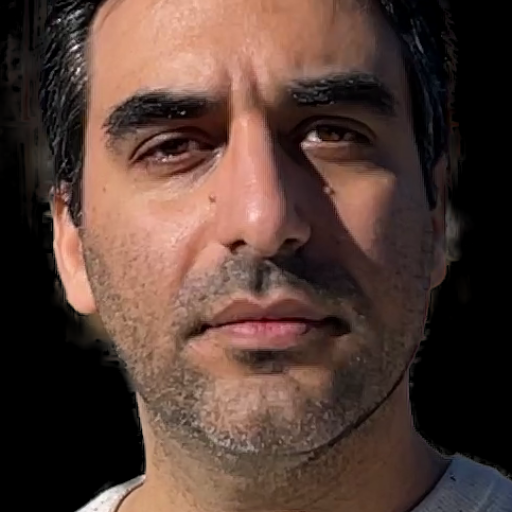}\\
 \caption{Input}
\end{subfigure} \hfill %
\begin{subfigure}[b]{0.24\linewidth}
 \centering
 \includegraphics[width=\linewidth]{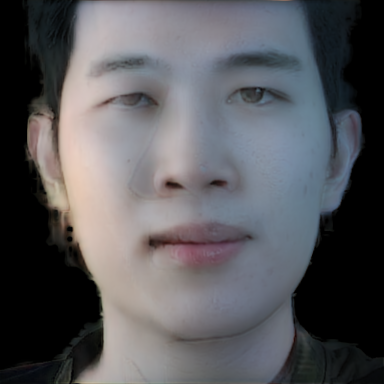}\\
 \includegraphics[width=\linewidth]{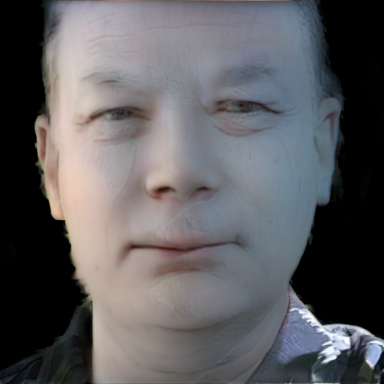}\\
 \includegraphics[width=\linewidth]{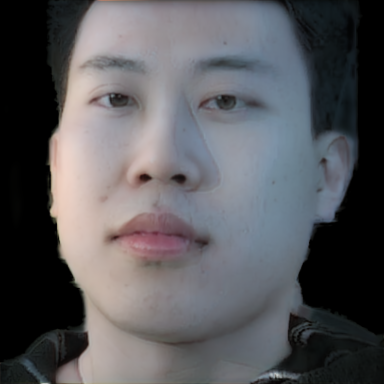}\\
 \includegraphics[width=\linewidth]{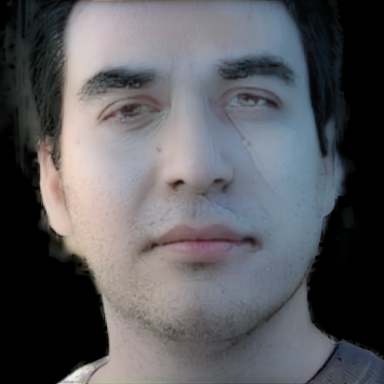}\\
 \caption{NVPR~\cite{zhang2021neuralvideo}}
\end{subfigure} \hfill %
\begin{subfigure}[b]{0.24\linewidth}
 \centering
 \includegraphics[width=\linewidth]{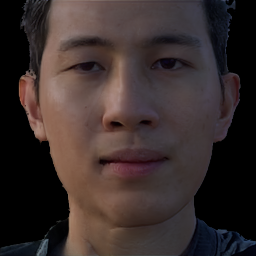}\\
 \includegraphics[width=\linewidth]{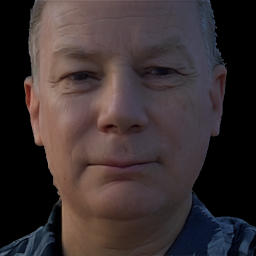}\\
 \includegraphics[width=\linewidth]{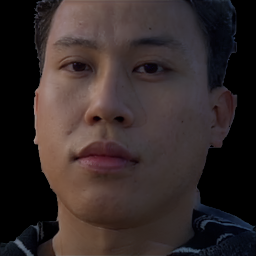}\\
 \includegraphics[width=\linewidth]{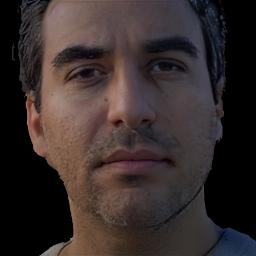}\\
 \caption{TR~\cite{pandey2021total}}
\end{subfigure} \hfill %
\begin{subfigure}[b]{0.24\linewidth}
 \centering
 \includegraphics[width=\linewidth]{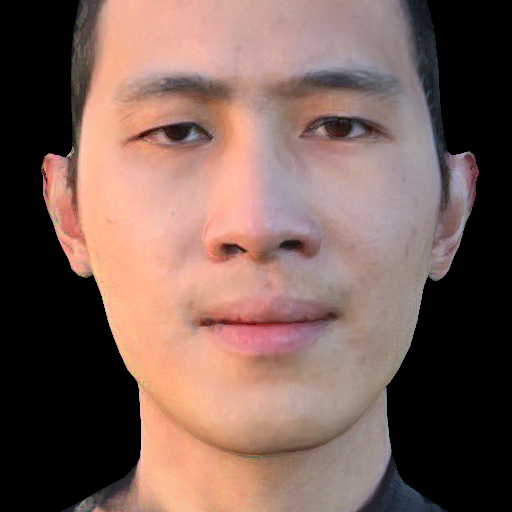}\\
 \includegraphics[width=\linewidth]{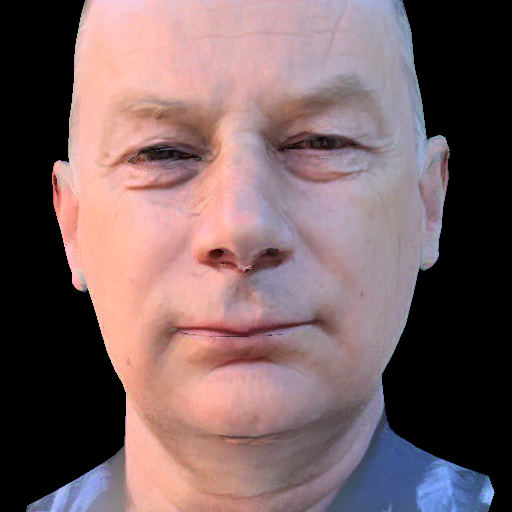}\\
 \includegraphics[width=\linewidth]{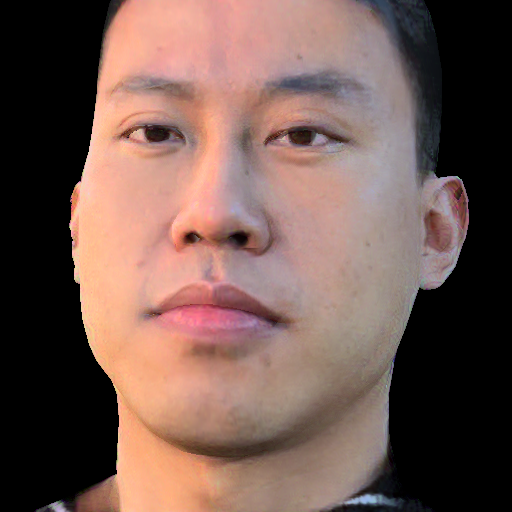}\\
 \includegraphics[width=\linewidth]{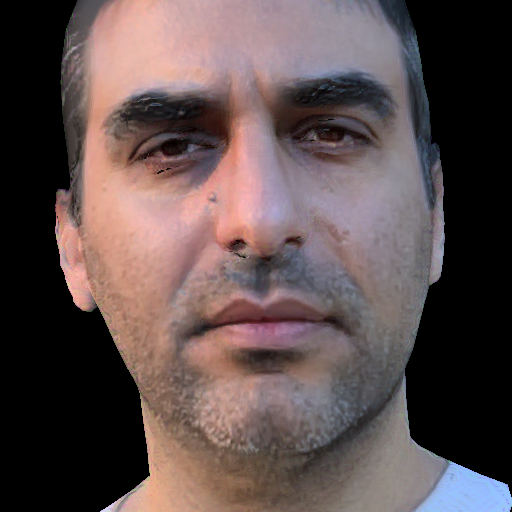}\\
 \caption{Ours}
\end{subfigure}
\end{center}
\caption{
Comparison with Neural Video Portrait Relighting and Total Relighting on the target lighting ``Hollywood''. Both NVPR and TR leverage face priors by training on large-scale light stage data. While being able to generalize to an arbitrary input, these methods can not model the individual skin reflectance properties. Both NVPR and TR also leave traces of visible shadow boundaries from the input images.
}
\label{fig:supp_tr_circus_2}
\end{figure}

\begin{figure}
\begin{center}

\centering
\begin{subfigure}[b]{0.24\linewidth}
 \centering
 \includegraphics[width=\linewidth]{fig/tr/input/aleks_2.png}\\
 \includegraphics[width=\linewidth]{fig/tr/input/dan_1.png}\\
 \includegraphics[width=\linewidth]{fig/tr/input/lars_3.png}\\
 \includegraphics[width=\linewidth]{fig/tr/input/cecilia_1.png}\\
 \caption{Input}
\end{subfigure} \hfill %
\begin{subfigure}[b]{0.24\linewidth}
 \centering
 \includegraphics[width=\linewidth]{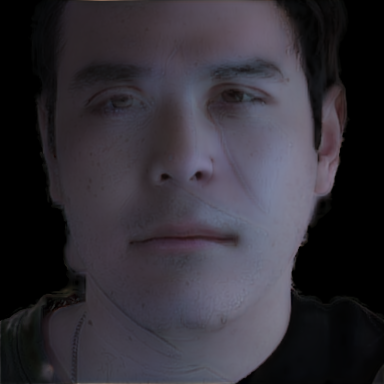}\\
 \includegraphics[width=\linewidth]{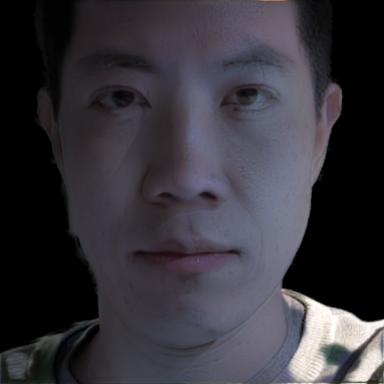}\\
 \includegraphics[width=\linewidth]{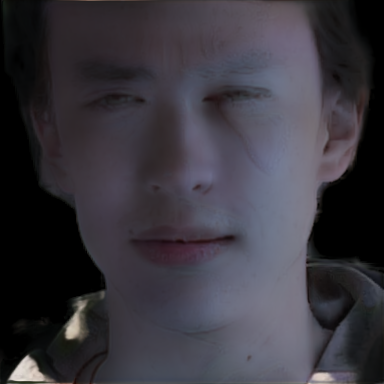}\\
 \includegraphics[width=\linewidth]{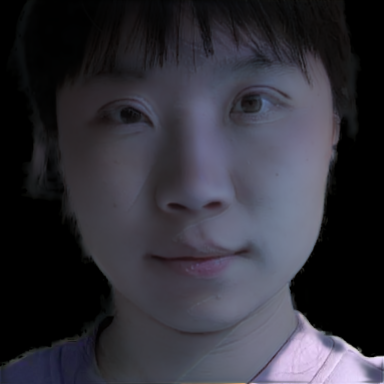}\\
 \caption{NVPR~\cite{zhang2021neuralvideo}}
\end{subfigure} \hfill %
\begin{subfigure}[b]{0.24\linewidth}
 \centering
 \includegraphics[width=\linewidth]{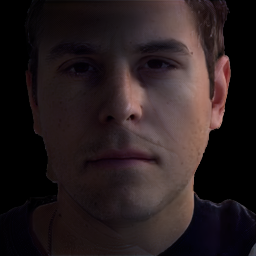}\\
 \includegraphics[width=\linewidth]{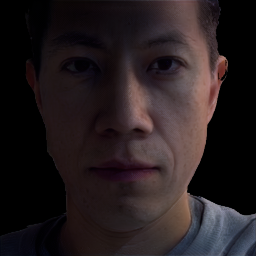}\\
 \includegraphics[width=\linewidth]{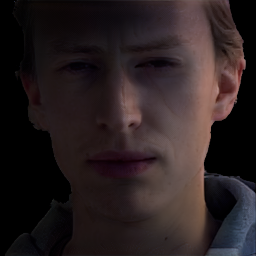}\\
 \includegraphics[width=\linewidth]{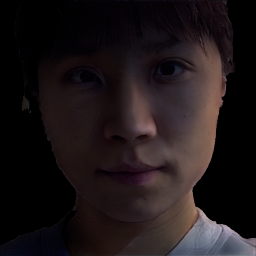}\\
 \caption{TR~\cite{pandey2021total}}
\end{subfigure} \hfill %
\begin{subfigure}[b]{0.24\linewidth}
 \centering
 \includegraphics[width=\linewidth]{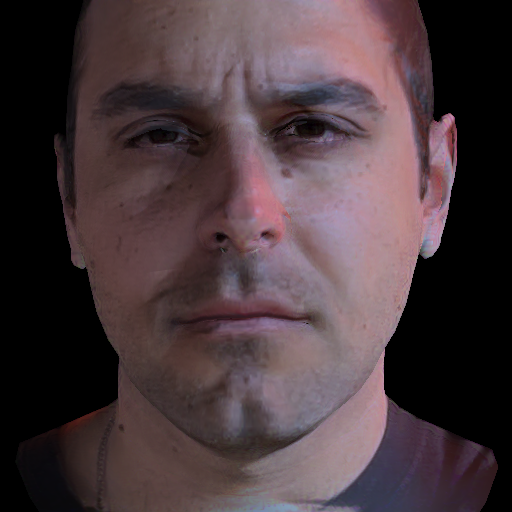}\\
 \includegraphics[width=\linewidth]{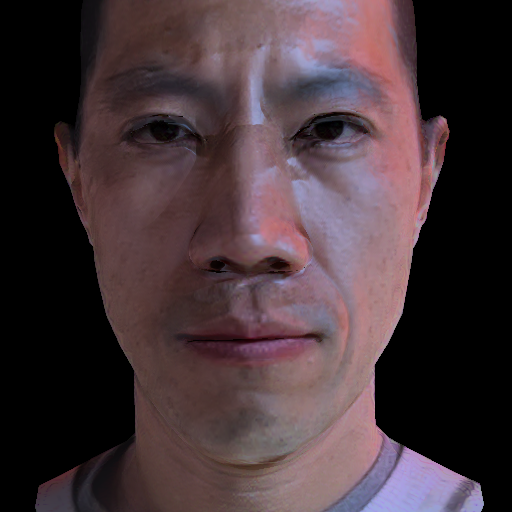}\\
 \includegraphics[width=\linewidth]{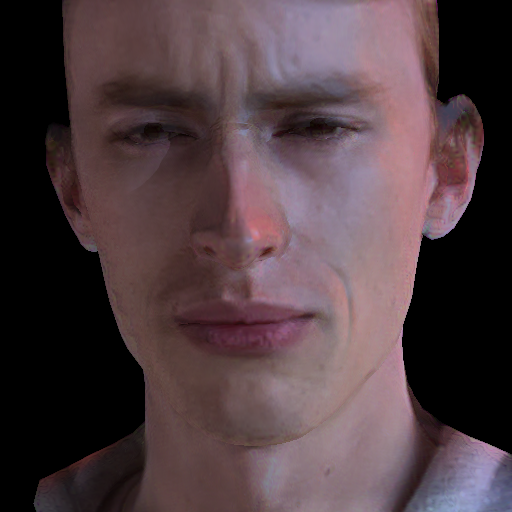}\\
 \includegraphics[width=\linewidth]{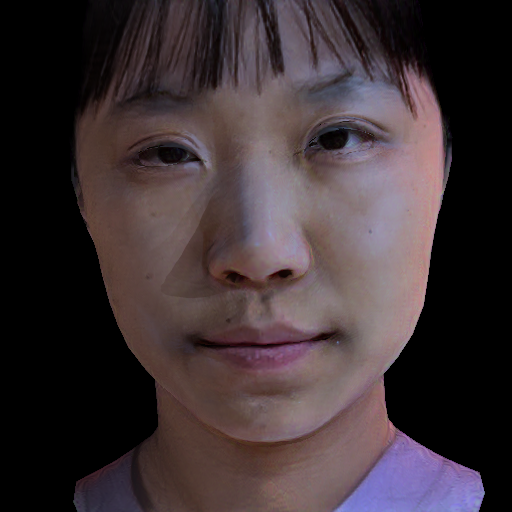}\\
 \caption{Ours}
\end{subfigure}
\end{center}
\caption{
Comparison with Neural Video Portrait Relighting and Total Relighting on the target lighting ``Queen Mary''. Both NVPR and TR leverage face priors by training on large-scale light stage data. While being able to generalize to an arbitrary input, these methods can not model the individual skin reflectance properties. Both NVPR and TR also leave traces of visible shadow boundaries from the input images.
}
\label{fig:supp_tr_qm_1}
\end{figure}

\begin{figure}
\begin{center}
\centering
\begin{subfigure}[b]{0.24\linewidth}
 \centering
 \includegraphics[width=\linewidth]{fig/tr/input/yifan_1.png}\\
 \includegraphics[width=\linewidth]{fig/tr/input/florian_2.png}\\
 \includegraphics[width=\linewidth]{fig/tr/input/xiuming_2.png}\\
 \includegraphics[width=\linewidth]{fig/tr/input/roy_1.png}\\
 \caption{Input}
\end{subfigure} \hfill %
\begin{subfigure}[b]{0.24\linewidth}
 \centering
 \includegraphics[width=\linewidth]{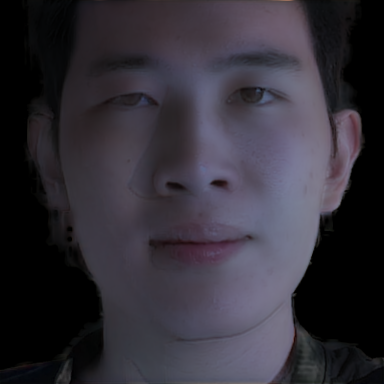}\\
 \includegraphics[width=\linewidth]{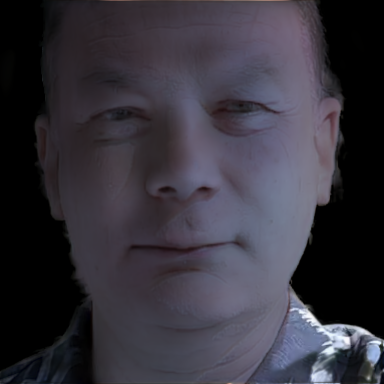}\\
 \includegraphics[width=\linewidth]{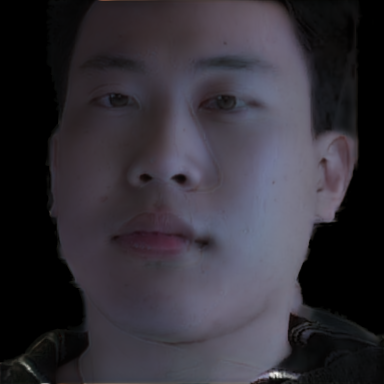}\\
 \includegraphics[width=\linewidth]{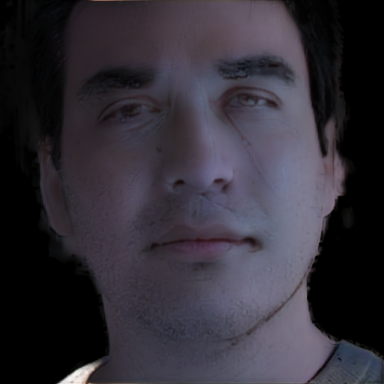}\\
 \caption{NVPR~\cite{zhang2021neuralvideo}}
\end{subfigure} \hfill %
\begin{subfigure}[b]{0.24\linewidth}
 \centering
 \includegraphics[width=\linewidth]{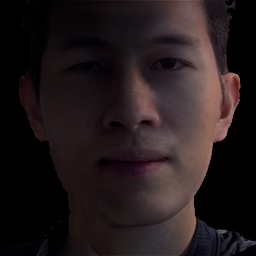}\\
 \includegraphics[width=\linewidth]{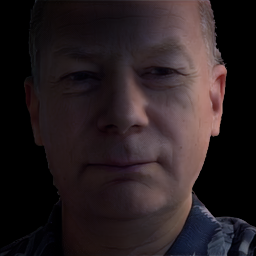}\\
 \includegraphics[width=\linewidth]{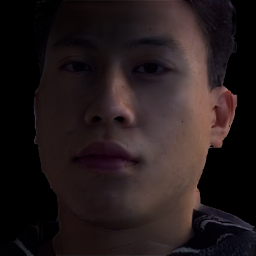}\\
 \includegraphics[width=\linewidth]{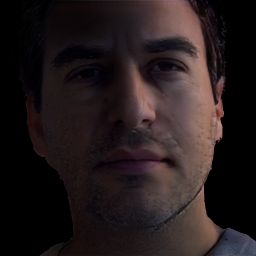}\\
 \caption{TR~\cite{pandey2021total}}
\end{subfigure} \hfill %
\begin{subfigure}[b]{0.24\linewidth}
 \centering
 \includegraphics[width=\linewidth]{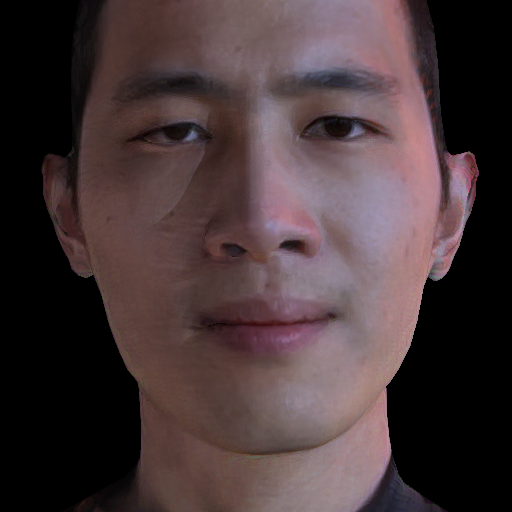}\\
 \includegraphics[width=\linewidth]{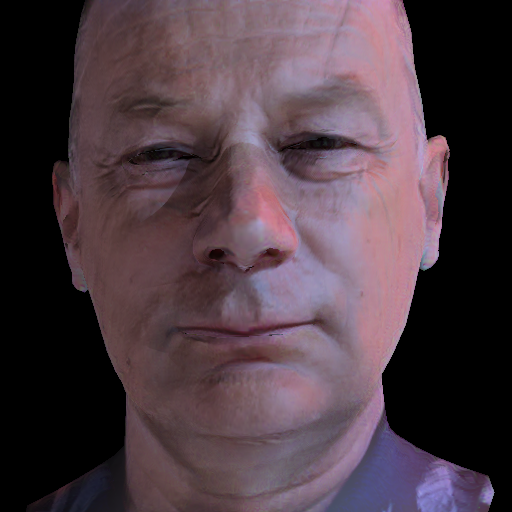}\\
 \includegraphics[width=\linewidth]{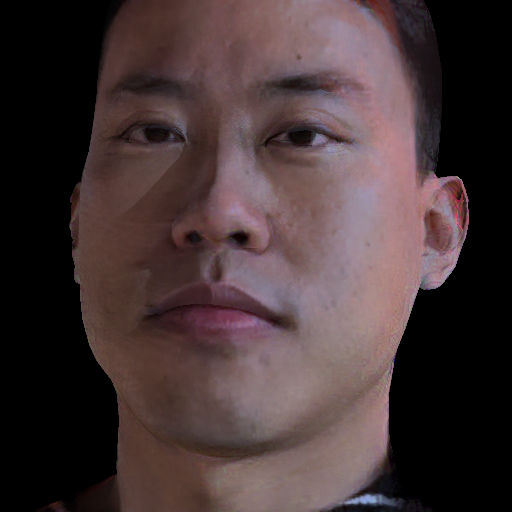}\\
 \includegraphics[width=\linewidth]{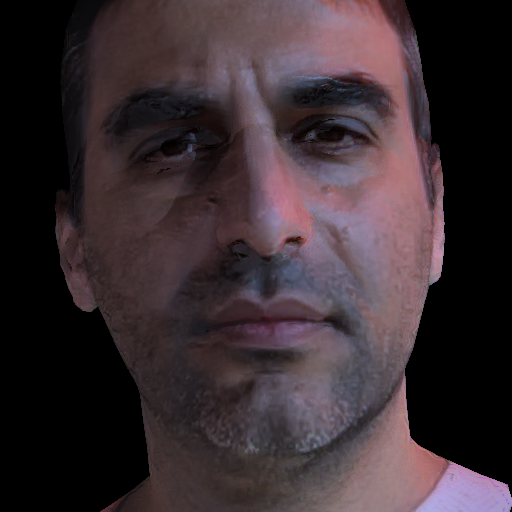}\\
 \caption{Ours}
\end{subfigure}
\end{center}
\caption{
Comparison with Neural Video Portrait Relighting and Total Relighting on the target lighting ``Queen Mary''. Both NVPR and TR leverage face priors by training on large-scale light stage data. While being able to generalize to an arbitrary input, these methods can not model the individual skin reflectance properties. Both NVPR and TR also leave traces of visible shadow boundaries from the input images.
}
\label{fig:supp_tr_qm_2}
\end{figure}
\begin{figure*}[t!]
\begin{center}
\includegraphics[width=\linewidth]{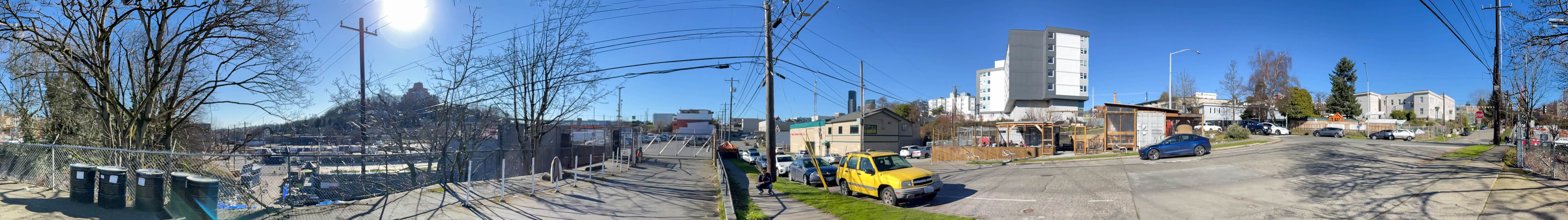}\\
\includegraphics[width=\linewidth]{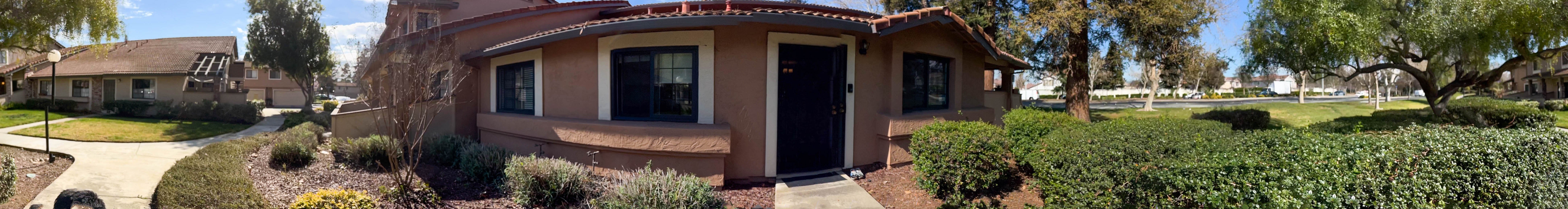}%
\end{center}
\vspace{-4mm}
\caption{
Example stitched panorama used for video background composite.
}
\label{fig:supp_envmap}
\end{figure*}
\begin{figure}
\begin{center}

\centering
\begin{subfigure}[t]{0.24\linewidth}
 \centering
 \includegraphics[width=\linewidth]{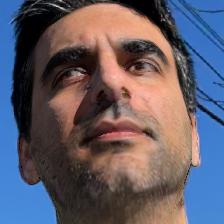} 
 \caption{Input}
\end{subfigure}
\begin{subfigure}[t]{0.24\linewidth}
 \centering
 \includegraphics[width=\linewidth]{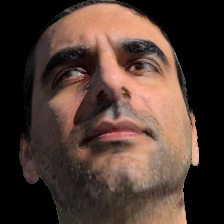} 
 \caption{{\scriptsize Reconstruction}}
\end{subfigure}
\begin{subfigure}[t]{0.24\linewidth}
 \centering
 \includegraphics[width=\linewidth]{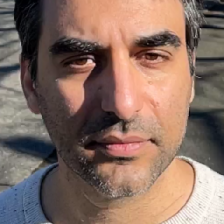} 
 \caption{Input}
\end{subfigure}
\begin{subfigure}[t]{0.24\linewidth}
 \centering
 \includegraphics[width=\linewidth]{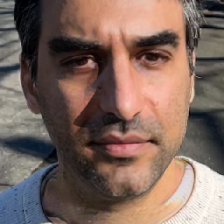}
 \caption{{\scriptsize Reconstruction}}
\end{subfigure}\\
\begin{subfigure}[t]{0.24\linewidth}
 \centering
 \includegraphics[width=\linewidth]{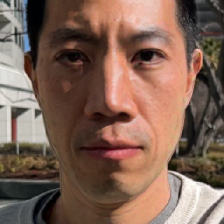} 
 \caption{Input}
\end{subfigure}
\begin{subfigure}[t]{0.24\linewidth}
 \centering
 \includegraphics[width=\linewidth]{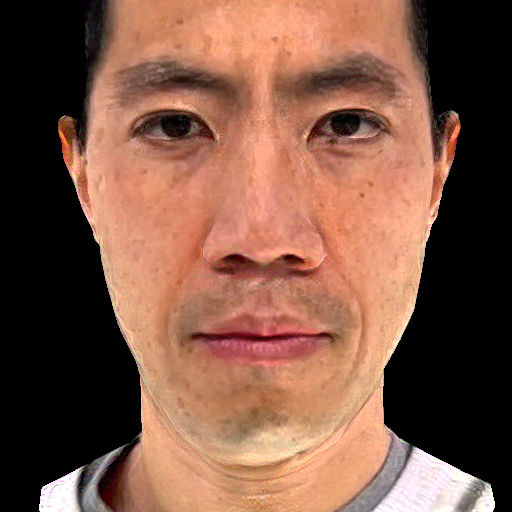} 
 \caption{Albedo}
\end{subfigure}
\begin{subfigure}[t]{0.24\linewidth}
 \centering
 \includegraphics[width=\linewidth]{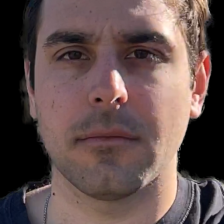} 
 \caption{Input}
\end{subfigure}
\begin{subfigure}[t]{0.24\linewidth}
 \centering
 \includegraphics[width=\linewidth]{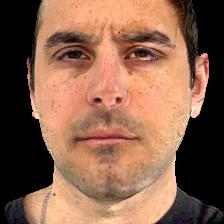} 
 \caption{Albedo}
\end{subfigure}\\
\begin{subfigure}[t]{0.24\linewidth}
 \centering
 \includegraphics[width=\linewidth]{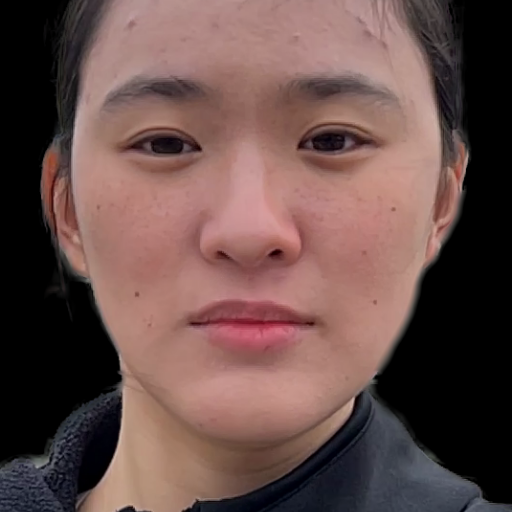} 
 \caption{Input}
\end{subfigure}
\begin{subfigure}[t]{0.24\linewidth}
 \centering
 \includegraphics[width=\linewidth]{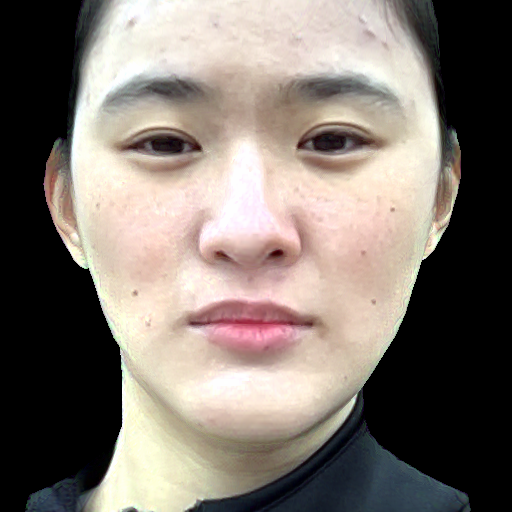} 
 \caption{Albedo}
\end{subfigure}
\begin{subfigure}[t]{0.24\linewidth}
 \centering
 \includegraphics[width=\linewidth]{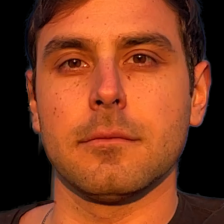} 
 \caption{Input}
\end{subfigure}
\begin{subfigure}[t]{0.24\linewidth}
 \centering
 \includegraphics[width=\linewidth]{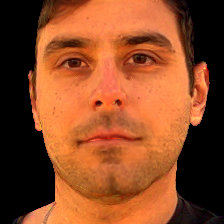}
 \caption{Albedo}
\end{subfigure}\\
\caption{\textbf{Limitations}. SunStage has limitations in its physical model and capture setup. (a, b) and (c, d) show reconstructions that fail to model hair and clothes. (e, f) and (g, h) illustrate the reconstructed albedo entangled with highlights around the chin region, which does not see lighting variations at capture time. (i, j) and (k, l) show failure cases when the capture lighting requirement breaks. The former is captured under a cloudy day and the latter is captured under a strongly tinted lighting condition. (k, l) shows the same identity as in (g, h) captured under different lighting conditions. The difference in the predicted albedo demonstrates the albedo-illuminant ambiguity, and the need for the assumption of mid-day (or otherwise known) sun color.}
\vspace{-3em}
\label{fig:limitation}
\end{center}

\end{figure}

\section{Applications}

\paragraph{OLAT}
To further validate the quality of the reconstructed geometry and the material properties, we simulate the One-Light-at-A-Time (OLAT) lighting setup typically seen in light stage captures \cite{debevec2000acquiring}.
Our results in Figure~\ref{fig:app-olats} show that we can plausibly recreate this challenging lighting setup, which typically exposes most errors in the estimated geometry and reflectance.

\paragraph{Relighting: Soften shadows}
To soften harsh shadows, we increase the size of the light source by applying a random offset $j$ to the (optimized) sun position $p_\text{sun}$. This offset can be interpreted as the radius of a virtual area light -- $j$ controls the size of the light and thus the softness of the shadow. We sample $n$ new sun positions, and average these $n$ renders to produce the resulting rendering with softened shadows.

\paragraph{Relighting: Lighting replacement}
We can use the learned properties to realistically render the subject with a new input environment map. The input environment map is downsampled to $16 \times 32$. To render, each pixel in the environment map is treated as a directional light source. The diffuse and specular contribution is calculated following Equation~6 and Equation~7 in the main paper.
% ~\ref{eq:dif_c5} and Equation~\ref{eq:sp_c6}. 

\paragraph{Relighting: Different time of day}
We show that we are able to simulate relit faces from an arbitrary time of day, including the fleeting ``golden hour'' and ``blue hour'' lighting that is favoured by many portrait photographers. To do so, we look up the correlated color temperature (in Kelvin) for different time of day, convert the color temperature into color matrices in the sRGB space, and use these to change the color of the sun in rendering.

\paragraph{View Synthesis}
We can change the (optimized) camera parameters to synthesize novel views of the subject. We can also render the subject with different amounts of perspective effects by changing the (optimized) camera focal length. In practice, we linearly scale the focal length and the subject distance to preserve the size of the face in the frame, as is done in a dolly zoom.

\section{Visualization Details}
\subsection{Compositing background}
We use different compositing methods to combine the rendered foreground subject and background for different applications.

\paragraph{Black background}
We use a black background (i.e. do not do compositing) for the One-Light-At-a-Time rendering (Figure~\ref{fig:app-olats}), to mimic the capture setup of a light stage. We also use a black background for more dramatic lighting setups that are similar to studio lighting, like the blue fill light shown in Figure~1b in the main paper.
% \ref{fig:teaser}b. 
We find that significant changes in color to the original scene's lighting tend to look unrealistic when composited onto the original background.

\paragraph{Original background}
Whenever possible, we use the original input image as the background in compositing. Note that the original background contains a portion of the hair that is not modeled physically, and thus does not respect changes in lighting, viewpoint, or other parameters. As such, the cases in which we can realistically composite onto the original background are limited, and only include shadow softening and subtle changes to lighting direction and magnitude.

\paragraph{Panorama background}
For the remainder of cases, when we would like the subject to remain in the original scene, but the lighting or viewpoint have changed significantly from the observed input frames, we instead composite the subject onto a panorama of the original scene. This panorama is automatically stitched from the input video frames (masking out the subject in each frame, i.e., $I_j \cdot (1 - I_\text{mask})$). See examples in Figure~\ref{fig:supp_envmap}.

\section{Discussions and Limitations}
\label{supp-limitation}
\paragraph{Physical model}
Our method inherits the limitations of existing morphable models that do not model hair, teeth, clothes, or accessories. Figure~\ref{fig:limitation} (a, b) shows a reconstruction that does not model the hair, and Figure~\ref{fig:limitation} (c, d) shows an example where the reconstruction fails to model the clothes.

\paragraph{Capture}
Our capture setup is not always comprehensive enough to model the full reflectance of the face. There are regions of the face that may not observe changes in lighting during the entire capture, like the bottom of the chin, which is often under shade. This causes ambiguity in our reconstruction, since the observed color can be explained by different combinations of albedo and lighting. Shown in Figure~\ref{fig:limitation} (e, f) and (g, h), this can result in highlights baked into the albedo.

Additionally, our optimization makes assumptions about the scene lighting: 1) the sun must be the dominant light source (i.e. the method does not work for a cloudy day capture), Figure~\ref{fig:limitation} (i, j) shows an example where the video captured under a cloudy day does not produce a reasonable albedo, as the face is always observed under shade, without any specular or shadow constraints. and 2) the sun's color temperature must be roughly in the range of 5500K-6500K (i.e. daylight around noon). Otherwise our reconstruction can not resolve the ambiguity between the illuminant and albedo. Figure~\ref{fig:limitation} (k, l) shows a video captured at golden hour, a strongly tinted lighting, which induces ambiguity in the recovered albedo. The result of the same identity captured under the required lighting condition is shown in Figure~\ref{fig:limitation} (g, h), which has a much more reasonable albedo reconstruction.

\clearpage

\end{document}